\documentclass{article} % For LaTeX2e
\usepackage[preprint]{colm2026_conference}
\usepackage{lipsum}
\usepackage{microtype}
\usepackage{xspace}
\usepackage{booktabs}
\usepackage{graphicx}
\usepackage{amsmath}
\usepackage[table]{xcolor}
\usepackage{algorithm}
\usepackage{algpseudocode}
\usepackage{amsmath, amssymb}
\usepackage{float}
\usepackage[most]{tcolorbox}   % For tcolorbox with enhanced features
\usepackage{xcolor}            % For defining colors
\usepackage{verbatim}          % For verbatim environment inside boxes

\definecolor{promptbg}{RGB}{245,245,245} % light gray background

\usepackage{tabularx}  % in preamble
\usepackage{booktabs}    % \toprule, \midrule, \bottomrule, \cmidrule
\usepackage{graphicx}    % \resizebox
\usepackage[table]{xcolor}
\usepackage{hyperref}
\usepackage{url}
\usepackage{microtype}
\usepackage{booktabs}
\usepackage{wrapfig}
\usepackage{subcaption}
\usepackage{placeins} % important
\usepackage{graphicx}
\makeatletter
\DeclareRobustCommand\onedot{\futurelet\@let@token\@onedot}
\def\@onedot{\ifx\@let@token.\else.\null\fi\xspace}
\def\eg{\emph{e.g}\onedot} 
\def\hero{\textit{\textbf{HERA}}\xspace}
\def\qwen{Qwen-3-14B\xspace}
\def\llama{Llama-3.1-8\xspace}
\def\rope{{RoPE}\xspace}

\usepackage{lineno}

\definecolor{darkblue}{rgb}{0, 0, 0.5}
\hypersetup{colorlinks=true, citecolor=darkblue, linkcolor=darkblue, urlcolor=darkblue}

\title{Experience as a Compass: Multi-agent RAG with Evolving Orchestration and Agent Prompts}

\author{
Sha Li \\
Virginia Tech \\
\texttt{shal@vt.edu}
\And
Naren Ramakrishnan \\
Virginia Tech \\
\texttt{naren@vt.edu}
}
\begin{document}

\ifcolmsubmission
\linenumbers
\fi

\maketitle

\begin{abstract}
Multi-agent Retrieval-Augmented Generation (RAG), wherein each agent takes on a specific role, supports hard queries that
require multiple steps and sources, or complex reasoning. 
Existing approaches, however, rely on
static agent behaviors and fixed orchestration strategies, leading to brittle performance on diverse, multi-hop tasks. We identify two key limitations: the lack of continuously adaptive orchestration mechanisms and the absence of behavior-level learning for individual agents. To this end, we propose \hero, a hierarchical framework that jointly evolves multi-agent orchestration and role-specific agent prompts. At the global level, \hero optimizes query-specific agent topologies through reward-guided sampling and experience accumulation. At the local level, Role-Aware Prompt Evolution refines agent behaviors via credit assignment and dual-axes adaptation along operational and behavioral principles, enabling targeted, role-conditioned improvements. On six knowledge-intensive benchmarks, \hero achieves an average improvement of 38.69\% over recent baselines while maintaining robust generalization and token efficiency. Topological analyses reveal emergent self-organization, where sparse exploration yields compact, high-utility multi-agent networks, demonstrating both efficient coordination and robust reasoning.

\end{abstract}
\section{Introduction}
\label{sec:intro}
\vspace{-2pt}
Recent advances in Large Language Model (LLM)-based Multi-Agent Systems \citep{li2024survey} have enhanced Retrieval-Augmented Generation (RAG) \citep{lewis2020retrieval, gao2023retrieval, jiang2023active} through  collaborating specialized agents \citep{nguyen2025ma, liu2025hm, qian2024scaling, chen2025improving}, extending  tool use \citep{zhuang2023toolchain, schick2023toolformer, jin2025search} and memory integration \citep{hu2025memory, gutierrez2025rag, yang2025improving}. These systems are promising for complex and long-context reasoning, where effective coordination between retrieval and reasoning is critical. 

Despite their promise, existing multi-agent RAG frameworks are fundamentally constrained by static or weakly adaptive orchestration. Most prior work relies on fixed or sequential pipelines \citep{gamage2024multi, liu2025hm, yao2022react, qian2024scaling, chen2025improving}, which fails to accommodate query-dependent variation in reasoning and retrieval complexity, resulting in insufficient evidence collection or unnecessary retrieval overhead. Besides, these systems are highly susceptible to error propagation \citep{shen2025understanding, pei2025scope}, where mistakes compound across multi-turn interactions and long-horizon agent coordination. Since these errors are inherently compositional, reliance on global supervision signals (\eg, final answer correctness) \citep{chen2025improving, song2025r1old, song2025r1} provides limited attribution to specific failure sources, hindering targeted improvements. Moreover, mainstream training-based optimization remains costly and inflexible, typically requiring large trajectory datasets \citep{zhu2024information, chen2025improving, chen2025mao, zhu2024atm, jin2025search}, which constrains scalability and limits continual adaptation in dynamic knowledge environments \citep{tang2025adapting}. As query distributions and underlying corpora evolve, fixed reasoning and coordination strategies \citep{jeong2024adaptive} lead to stagnant performance and brittle generalization to novel queries and retrieval challenges.

To address these limitations, we argue that, an effective multi-agent RAG system must (i) adapt its coordination strategies to query complexity, (ii) enable targeted and role-specific agent improvement, (iii) mitigate error propagation, and (iv) support continual adaptation without costly retraining. To this end, we propose \hero, a \underline{\textbf{H}}ierarchical \underline{\textbf{E}}volvution \underline{\textbf{RA}}G framework that achieves these goals by \textit{distilling execution experience as in-context knowledge} and \textit{reshaping agent behaviors by prompt evolution} without parameter updates. 

\hero is organized as a three-layer hierarchy that jointly evolves global orchestration strategies and local agent behaviors using accumulated experiential and reflective knowledge (Figure .\ref{fig:hera}). At the top, a centralized \textit{orchestrator} generates query-specific execution plans in a single step. This holistic orchestration \citep{ke2026mas} enables the system to reason globally about the coordination of retrieval and reasoning, producing plans tailored to the complexity and requirements of each query. At the bottom, \textit{specialized agents} execute subtasks grounded in interaction with others. \hero employs role-aware prompt optimization to drive targeted improvements for each agent’s unique responsibilities, avoiding generic updates and mitigating errors arising from misaligned multi-agent behaviors. Bridging these layers, the \textit{experience library} captures reflective insights from both successful and failed execution trajectories. These insights provide diagnostic information \citep{agrawal2025gepa}, enabling semantic credit assignment for global learning and guiding agent-level refinement. Across \hero, continuously evolving experience acts as a compass: it informs the orchestrator to improve high-level coordination strategies and drives progressive, role-specific enhancement of agent behavior, improving robustness and reducing cascading errors.

Our contributions are summarized as follow: (i) We introduce \hero, a hierarchical framework enabling dual-level evolution - experience-guided orchestration and role-aware prompt adaptation, shifting policy optimization from parameter updates to experience-driven contextual adaptation. (ii) We design topology-based metrics to quantitatively characterize the evolution of agent interaction topologies and track dynamics. (iii) Empirical results on six knowledge-intensive benchmarks demonstrate that \hero outperforms recent SOTA by an average of 38.69\% while maintaining token efficiency. 

\vspace{-5pt}
\section{Preliminary and Problem Statement}
\vspace{-2pt}
\label{sec:probelm_formulate}
We study the problem of dual-level evolution in multi-agent RAG for knowledge-intensive tasks. Let $\mathcal{Q}$ denote a distribution over queries, $\mathcal{Y}$ the answer space, and $\mathcal{D}$ a corpus (\eg Wikipedia) providing supporting information. We formalize a whole system as a tuple:
\[\mathcal{R}=\langle \mathcal{O},\mathcal{N}, \mathcal{L},  \mathcal{T}, \mathcal{E}, \mathcal{A}, \mathcal{S}, \Psi, \tau,  \mathcal{Q}, \mathcal{Y}, \mathcal{D} \rangle \] where $\mathcal{O}$ is a central orchestrator, $\mathcal{N}$ is a set of specialized agents, $\mathcal{E}$ is an experience library, $\mathcal{S}$ is the state space, and $\mathcal{A}$ is the action space. Each agent $\mathcal{N}_i$ is abstracted as a tuple $(\pi_i, \rho_i, \mathcal{T}_i)$, where $\pi_i \in \mathcal{L}$ is the underlying LLM, $\rho_i$ is a role-specific prompt, and $\mathcal{T}i$ denotes the tools available to the agent. The overall action space $\mathcal{A}$ includes both reasoning operations and tool invocations. The system evolves according to the transition dynamics $\Psi(s_{t+1} \mid s_t, a_t)$, and agent executions produce trajectories $\tau = (s_0, a_0, s_1, a_1, \dots, s_T)$.

Given a query $q$, the orchestrator produces a multi-agent interaction topology conditioning on the experience library: $\Gamma\sim\pi_{\mathcal{O}}(\cdot|q, \mathcal{E}, \mathcal{N})$, where $\Gamma$ specifies the subset of agents, their execution order (sequential or parallel), and dependency. Agents execute according to $\Gamma$, producing a trajectory $\tau$, and the final answer $y \in \mathcal{Y}$ is synthesized from the information aggregated along the trajectory, typically based on the terminal state $s_T$.

Our goal is to jointly optimize the orchestration policy $\pi_{\mathcal{O}}$ and role-specific prompts ${\rho_i}$ while keeping the underlying LLM parameters ${\pi_i}$ frozen.

\vspace{-7pt}
\section{\hero: \texorpdfstring{\underline{H}ierarchical \underline{E}volution of Multi-agent \underline{RA}G}{Hierarchical Evolution of Multi-agent RAG}}
\label{sec:hero}
\vspace{-8pt}
\begin{figure}[h]
        \centering
        \includegraphics[width=1\linewidth]{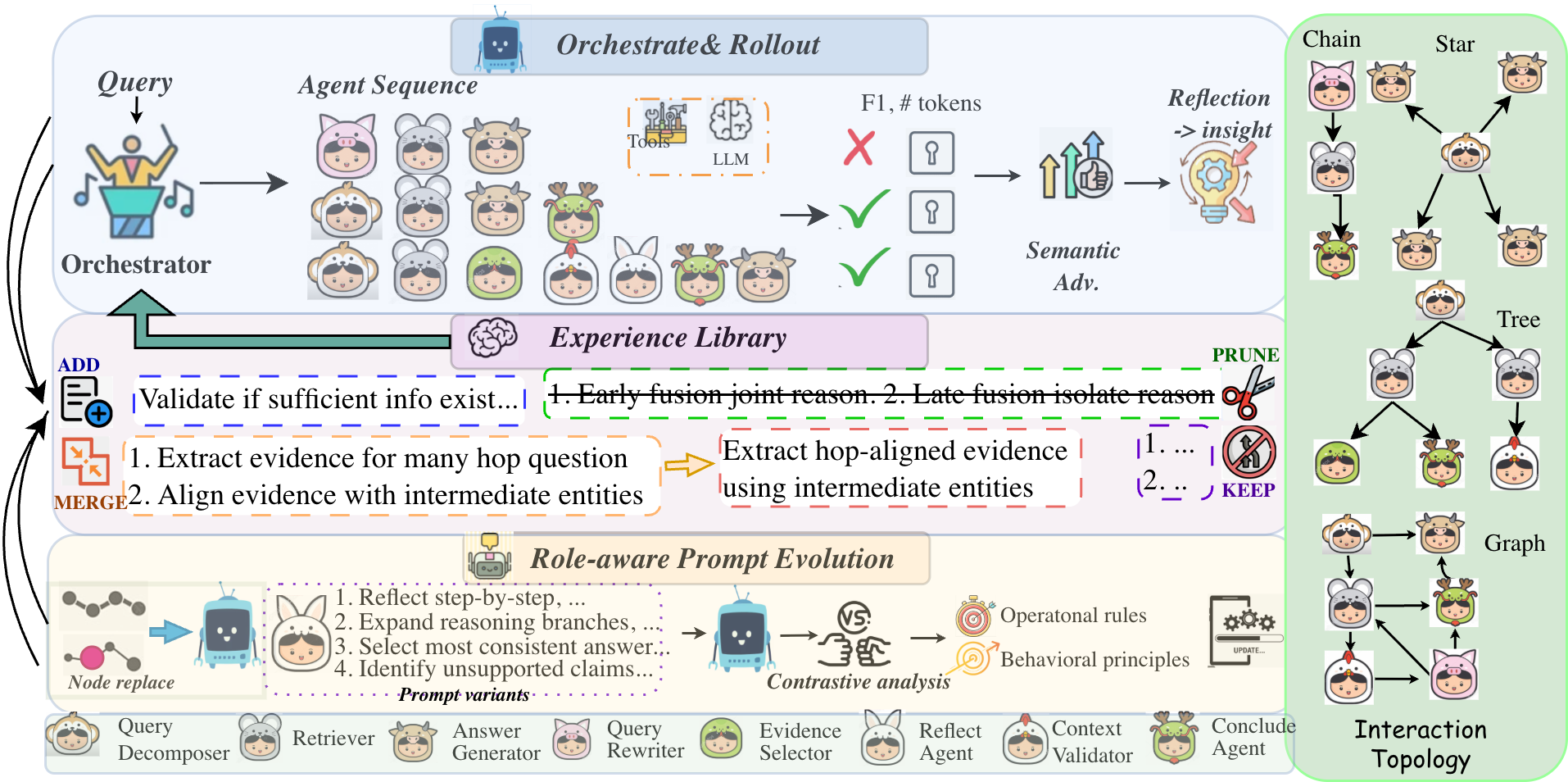}
        \caption{Overview of \hero. A hierarchical framework that jointly evolves orchestration strategies, the experience library, and agent prompts. }
        \label{fig:hera}
        \vspace{-20pt}
\end{figure}

\hero jointly evolves global orchestration and agent prompts through experience and reflection. As shown in Figure \ref{fig:hera}, it comprises three layers, each corresponding to a different level of optimization: the \textit{global orchestrator} optimizes high-level coordination strategies, the \textit{experience library} accumulates and refines trajectory-level knowledge to guide learning, and the \textit{execution agents} optimize role-specific performance through prompt evolution, per their operational roles and behavioral principles. 

\subsection{Orchestrator: Structure-Level Policy Optimization}
\label{sec:orchestrator_opt}

The orchestrator's optimization is inspired by Group Relative Policy Optimization (GRPO) \citep{shao2024deepseekmath, cai2025training}, which updates a policy by comparing sampled actions within a group. Unlike the training-free GRPO \citep{cai2025training}, we extend the idea beyond token-level generation, lifting optimization \textit{from token generation to cooperation topology}. Operating over a structured action space of agent topologies $\Gamma$, the orchestrator performs group-based optimization at the level of global strategy. Given a query $q$, it samples a group of candidate agent sequences. Each sequence is executed by invoking the corresponding agents, producing a trajectory $\tau_{i}$, which is subsequently evaluated to obtain a reward. Instead of relying on scalar advantages, we adopt a hierarchical comparison: trajectories are first ranked by downstream task performance (\eg, F1 score), and then by efficiency, measured as the total number of input and output tokens consumed across all agents. The orchestrator is then prompted to articulate the reasons for relative success and failure across the group, producing a set of natural language insights $\mathcal{I} = \pi_{\mathcal{O}}(\{\tau_i\}_{i=1}^G)$. To improve diagnostic precision, we focus on groups explicitly containing both successful and failed.
trajectories. 
 which serve as group-relative semantic advantages. These semantic advantages encode structured knowledge about effective reasoning strategies, agent interactions, and failure modes, effectively replacing numerical gradients with interpretable and compositional signals. As such, the orchestrator governs high-level structure and coordination (macro) instead of the fine-grained textual outputs (micro). These insights, curated through selective consolidation, form the core of the experience library (\S \ref{sec:exlib}).  

\vspace{-13pt}
\subsection{Experience Library}
\vspace{-6pt}
\label{sec:exlib}
Storing distilled experience as retrievable in-context demonstrations has been shown to effectively shape future behavior \citep{zhao2024expel, ouyang2025reasoningbank, xu2025mem}. Building on this principle, \hero employs an Experience Library $\mathcal{E}$, which accumulates semantic advantages derived from reflective comparisons of successful and failed trajectories, serving as both an episodic memory and an optimization medium for the orchestrator.

$\mathcal{E}$ is charatcerized with a \textit{Profile–Insight–Utility} structure $(c, z, u)$, where $c$ denotes the \textit{query characteristics or types}, $z$ is the natural language \textit{insight}, and $u$ represents the \textit{utility}, measuring the empirical success rate of $z$ - that is, how often the insight is successfully applied in subsequent agent orchestration. $\mathcal{E}$ is updated online through consolidation operations: \textcolor{blue}{ADD} inserts insights that are distinct from any existing entries; \textcolor{orange}{MERGE} combines semantically similar or complementary entries, 
\textcolor{green}{PRUNE} removes conflicting or low-utility insights to maintain generalizable insights while preventing uncontrolled growth, and 
\textcolor{purple}{KEEP} leaves $\mathcal{E}$ unchanged. 

\textbf{Experience as a Prior.} Given a query $q$, the orchestrator first analyzes its characteristics and retrieves a set of relevant insights from $\mathcal{E}$. Retrieval balances two objectives: maximizing empirical \textit{utility} $u$, which favors insights that have historically led to successful coordination, and maintaining \textit{diversity}, achieved by selecting insights that are semantically distinct from those already chosen and that have lower prior usage frequency. The retrieved insights inform the orchestrator’s selection and sequencing of agents, guiding multi-step execution toward high-performing strategies. By combining high-utility and diverse insights, the orchestrator biases execution to maximize task accuracy while minimizing token consumption. In this manner, $\mathcal{E}$ serves as a structured, experience-driven prior, enabling contextual policy improvement and without modifying the underlying LLM.

\vspace{-6pt}
\subsection{Theoretical Interpretation}
\vspace{-6pt}
The \hero orchestrator admits a natural \textbf{Expectation-Maximization (EM)} interpretation over discrete reasoning programs. In the \textbf{E-step}, trajectories are sampled under the current policy, approximating the posterior over high-reward reasoning topologies. In the \textbf{M-step}, semantic insights are extracted and consolidated into the experience library, updating a latent representation of the policy. Further, using a frozen LLM introduces implicit regularization, analogous to KL-constrained policy optimization. As updates occur via contextual augmentation instead of parameter changes, the induced policy remains close to the original model distribution. This formulation highlights that \hero performs energy-based reweighting of candidates, ensuring stable and consistent improvement. We provide detailed theoretical analysis in Appendix \ref{app:theory}.

\vspace{-6pt}
\subsection{Role-aware Prompt Evolution (\rope)}
\vspace{-6pt}
\label{sec: rapo}
Globally strategy cannot prevent \textbf{corrective} (agents ignore actionable error signals and repeat mistakes) or \textbf{enhancement failures} (agents persist with suboptimal strategies despite successful execution) \citep{zhang2025agent} in multi-agent systems, motivating our\textit{ Role-aware Prompt Evolution} (\rope).

\vspace{-6pt}
\subsubsection{Prompt Evolution}
\label{sec:agent_projected}
\vspace{-4pt}
A key challenge in multi-agent systems is attributing global trajectory outcomes to individual agents, especially when inter-agent dependencies obscure direct credit assignment. \rope addresses this by targeting agents identified as \textit{underperforming} by the orchestrator, who evaluates complete trajectories. Formally, an agent is considered underperforming if the orchestrator identifies it as the primary contributor to the trajectory’s failure.

To support systematic improvement, we maintain a buffer for each agent containing its recent failed trajectories. This buffer enables RoPE to analyze recurring errors, identify consistent failure patterns, and provide contextual evidence for targeted prompt refinement. Given an underperforming agent $\mathcal{N}_{i}$ at time $t$, we retrieve its recent failed trajectories, then we generate prompt variants, and reexecutet the original whole agent trajectories with these prompt variants. Then \rope performs contrastive analysis to extract actionable improvements along two complementary dimensions: (i) \textit{operational rules ($\Delta\rho_i^{op}$)}: short-term corrective behaviors derived from recent failures, and (ii) \textit{behavioral principles ($\Delta\rho_i^{bp}$)}: long-term strategies distilled from patterns across multiple trajectories. Formally, the prompt update can be decomposed as:
\vspace{-5pt}
\[
\Delta\rho_{i}=\Delta\rho_{i}^{op}+\Delta\rho_{i}^{bp}
\]
\vspace{-2pt}
where $\Delta\rho_{i}^{op}$ addresses \textit{immediate failure correction} and $\Delta\rho_{i}^{bp}$ promotes \textit{strategic generalization}. Prompt variants are generated along behavioral axes such as thoroughness, risk sensitivity, error correction and heuristic injection, enabling structured exploration to optimize accuracy, robustness, and computational cost (see Appendix~\ref{app:promptt} for prompt details).
\vspace{-5pt}

\subsubsection{Prompt Consolidation}
\vspace{-2pt}
Following variant evaluation, selected updates are integrated into the agent’s prompt via prompt consolidation, which prunes redundant or low-utility instructions to maintain a compact and coherent representation. Formally, this can be expressed as a projection:
\vspace{-2pt}
\[
\rho_{i}^{t+1} = \Pi_{\mathcal{C}}\left(\rho_{i}^{t} \oplus \Delta \rho_{i}\right)
\]

where $\mathcal{C}$ represents constraints on prompt length and coherence. With the underlying LLM fixed, prompt evolution induces a controlled shift in the agent’s effective policy:
\[
\pi_{\mathcal{N}_{i}^{new}}(y|x)\propto\pi_{\mathcal{N}_{i}^{base}}(y|x)\cdot exp(f{\rho_{i}}(x, y))
\]
where $x, y$ denote the input and output context of an agent respectively, and $\pi_{\mathcal{N}_{i}^{new}}$ represents its conditional generation policy. The prompt-induced term $f{\rho_{i}}(x, y)$
biases this mapping by favoring outputs 
$y$ that better align with the updated role-specific guidance. This induces an implicit regularization effect, analogous to KL constraints, ensuring stable and consistent policy updates.
\vspace{-8pt}
\subsection{Topology Mutation}
\vspace{-8pt}
While prompt evolution improves agent-level behavior, persistent failures often indicate structural deficiencies in the coordination topology. To address this, \hero performs topology mutation when trajectories consistently fail (\eg, F1 = 0). Upon identification by the orchestrator, \hero explores alternative structures by either replacing the failed agent $\mathcal{N}_{i}$ with an alternative $\mathcal{N}_{i}^{'}$ or augmenting the topology with additional agents, yielding candidate topologies $\Gamma^{'}$. These candidates are then incorporated into the same gradient-free GRPO optimization loop. This process enables the co-evolution of coordination structure and agent behaviors.

\vspace{-8pt}
\section{Experiments}
\label{sec:exp}
\vspace{-10pt}
We aim to answer the following questions: \textbf{RQ1 (Effectiveness)}: How does \hero’s overall performance compare to state-of-the-art (SOTA) baselines? (\S \ref{sec:overall_perf}) \textbf{RQ2 (Component Contribution)}:  What is the individual contribution of the key components? (\S\ref{sec:ablation}) \textbf{RQ3 (Efficiency)}: Does improved performance, if any, come at the cost of increased token consumption?(\S\ref{sec:token_usage}) \textbf{RQ4 (Topology Evolution)}: How do agent coordination topologies evolve as the system accumulates experience and updates agent prompts? (\S\ref{sec:topology_analysis})

\textbf{Tasks and Datasets.}We evaluate \hero on multi-hop QA benchmarks 2WikiMultiHopQA (2WikiQA) \citep{ho-etal-2020-constructing}, HotpotQA \citep{yang2018hotpotqa}, MusiQue \citep{trivedi2021musique} and AmbigQA \citep{min-etal-2020-ambigqa} for ambiguous QA. For out-of-distribution (OOD) evaluation, we use Bamboogle \citep{press-etal-2023-measuring} and HoVer \citep{jiang2020hover} for multi-hop QA and fact verification respectively. Evaluation metrics involve Exact Match (EM), accuracy, and F1 score for QA tasks, and accuracy for fact verification.

\textbf{Baselines.}
We compare \hero against:
(1) \textbf{Direct inference and chain-of-thought (CoT) without RAG} across different LLM families and sizes;
(2) \textbf{Single-turn RAG} with different LLMs as answer generators; (3) \textbf{Iterative RAG}: IterDRAG \citep{yue2024inference}, Plan-RAG \citep{verma2024plan}
Search-o1 \citep{li2025search}, Search-R1 \citep{jin2025search}, IRCoT \citep{trivedi-etal-2023-interleaving}, SELF-RAG \citep{asai2023self}, CORAG \citep{wang2025chain}; and (4) \textbf{Agentic RAG}, including InstructRAG, R1-Searcher \citep{song2025r1old},  DeepResearcher \citep{zheng-etal-2025-deepresearcher}, MMOA-RAG \citep{chen2025mao}, AceSearcher \citep{xu2025acesearcher} , and MAO-ARAG \citep{chen2025mao}, ExSearch \citep{shi2025iterative}.
Model and implementation details are provided in Appendix~\ref{app:baselin_model}.

%, including Qwen-3-4B, 8B, and 14B \citep{yang2025qwen3}, Qwen2.5-Instruct-7B \citep{hui2024qwen2}, LLaMA3.1-8B \citep{grattafiori2024llama}, and GPT-4o-mini \citep{hurst2024gpt};

\textbf{Implementation Details.} We consider core agents commonly employed in RAG: Query Decomposer, Retriever, Answer Generator, Query Rewriter, Evidence Selector, Context Validator, Reflect Agent, and Conclude Agent. We use Wikipedia as the corpus, returning top-\emph{5} passages per step for multi-step retrieval and top-\emph{10} for single-turn methods via the BGE retriever \citep{xiao2024c}. We use \llama \citep{grattafiori2024llama} and \qwen \citep{yang2025qwen3} as the backbone LLMs for the orchestrator and GPT-4o-mini for agents. All backbones remain frozen during learning and inference. To highlight data efficiency, we first use GPT-4o to annotate queries from 2WikiQA, HotpotQA, AmbigQA, and MusiQue along \textbf{reasoning type} (bridge, intersection, comparison, temporal multi-hop, ambiguous) and \textbf{complexity} (easy, medium, hard). Based on these annotations, we perform stratified, difficulty-aware sampling to construct a curated training set that ensures balanced coverage across reasoning categories while preserving distributional diversity (statistics in Appendix~\ref{app:training_set}). Generative sampling uses temperature 0.9 during group rollouts, and $0.0 / 0.3$ for in-distribution and OOD evaluation, respectively.  (Pseudocode for \hero is provided in Appendix~\ref{app:pseducode}).

\vspace{-8pt}
\section{Results and Analysis}
\label{sec:res}
\vspace{-5pt}
\subsection{Overall Performance}
\vspace{-3pt}
\label{sec:overall_perf}
\begin{table}[ht]
\centering
\scriptsize
\setlength{\tabcolsep}{2pt}
\vspace{-5pt}
\caption{Comparison of \hero with baselines. Best results are \textbf{bold}, second-best are \underline{underlined}. ``-'' indicates methods not designed for or evaluated on the task.} 
\label{tab:main_exp_res}
\vspace{-5pt}
\resizebox{\linewidth}{!}{
\begin{tabular}{l|ccc|ccc|ccc|ccc|ccc|c}
\toprule
\textbf{Dataset} 
& \multicolumn{3}{c|}{HotpotQA} 
& \multicolumn{3}{c|}{2WikiQA} 
& \multicolumn{3}{c|}{MusiQue} 
& \multicolumn{3}{c|}{AmbigQA} 
& \multicolumn{3}{c|}{Bamboogle} 
& HoVer \\
\cmidrule(r){2-4} \cmidrule(r){5-7} \cmidrule(r){8-10} 
\cmidrule(r){11-13} \cmidrule(r){14-16} \cmidrule(r){17-17}
\textbf{Metrics} 
& Acc & EM & F1 
& Acc & EM & F1 
& Acc & EM & F1 
& Acc & EM & F1 
& Acc & EM & F1 
& Acc \\
\midrule
\multicolumn{17}{l}{\textbf{\textit{Single-turn Retrieval}}} \\
\midrule
Qwen2.5-14B-Instruct & 43.53 & 35.72 & 46.14 & 26.64 & 26.25 & 31.85 & 9.23 & 7.00 & 13.75 & 39.78 & 36.65 & 50.23 & 27.20 & 34.70 & 34.35 & 45.50  \\
Qwen3-14B & 44.47 & 36.25 & 47.56 & 31.57 & 31.10 & 37.79 & 13.16 & 9.99 & 15.72 & 38.47 & 36.10 & 50.91 & 31.54 & 29.60 & 38.31 & 49.05 \\
Qwen3-8B & 41.35 & 33.70 & 45.10 & 26.37 & 26.10 & 32.12 & 11.49 & 8.71 & 13.62 & 36.02 & 33.80 & 47.70 & 23.87 & 22.40 & 30.28 & 50.15 \\
GPT-4o-mini & 46.30 & 37.74 & 49.30 & 28.42 & 28.41 & 35.11 & 16.63 & 13.94 & 19.60 & 37.40 & 33.80 & 47.99 & 38.36 & 36.00 & 42.81 & 47.95 \\
Llama3-3.1-8B-instruct & 33.16 & 27.02 & 35.14 & 20.40 & 20.35 & 25.11 & 8.43 & 6.43 & 12.61 & 18.03 & 18.96 & 24.60 & 20.46 & 19.20 & 27.24 & 49.90  \\
\midrule
\multicolumn{17}{l}{\textbf{\textit{Advanced RAG}}} \\
\midrule
Search-r1-7B & 51.11 & 41.94 & 54.18 & 48.89 & 44.69 & 49.20 & 20.45 & 18.58 & 25.73 & 47.32 & \underline{45.88} & \underline{65.54} & 35.79 & 32.64 & 43.93 & 47.76 \\
Iter-RetGen (Qwen2.5-7B) & 20.37 & 16.71 & 24.59 & 47.71 & 43.54 & 49.12 & 19.89 & 16.79 & 21.04 & --- & --- & --- & 35.33 & 34.12 & 40.65 & 51.29 \\
Plan-RAG 8b & 29.95 & 25.96 & 37.51 & 30.53 & 27.38 & 31.68 & 13.08 & 9.90 & 13.34 & --- & --- & --- & 30.01 & 23.57 & 32.12 & 50.33 \\
SELF-RAG (llama2-7B) & 23.85 & 13.64 & 25.07 & 40.82 & 37.30 & 42.02 & 8.47 & 7.16 & 14.16 & 28.51 & 24.03 & 39.62 & 25.80 & 22.66 & 30.05 & 40.52 \\
R1-Searcher 7B & 41.15 & 32.54 & 50.69 & 40.11 & 35.38 & 45.19 & 18.21 & 13.51 & 19.19 & --- & --- & --- & 41.00 & 38.85 & 46.76 & 51.56 \\
ExSearch (Qwen2.5-7B) & \underline{55.04} & 45.11 & 58.22 & 44.83 & 44.03 & 51.56 & 22.73 & 19.58 & 27.10 & -- & --- & -- & \textbf{56.12} & \textbf{51.27} & \textbf{63.80} & \underline{66.62} \\
DeepResearcher-7B & 40.65 & 35.23 & 58.86 & 49.40 & \underline{54.35} & \underline{64.05} & 24.53 & 18.56 & 25.57 & \underline{49.74} & 42.23 & 51.50 & 37.42 & 34.20 & 44.40 & 58.19 \\
IRCoT (GPT-4o-mini) & 40.39 & 33.14 & 54.93 & 38.73 & 35.50 & 42.01 & 19.11 & 18.23 & 25.00 & --- & --- & -- & 34.88 & 30.73 & 33.97 & 55.37 \\
InstructRAG-8B & 34.40 & 33.15 & 34.83 & 36.98 & 35.45 & 40.72 & 16.11 & 14.19 & 19.50 & 40.07 & 39.72 & 40.73 & 26.16 & 23.44 & 31.18 & 56.76 \\
CORAG-8B (Greedy) & 38.02 & 32.95 & 52.85 & 39.09 & 35.43 & 46.62 & 16.45 & 15.98 & 21.00 & 32.10 & 27.25 & 43.95 & 33.03 & 31.61 & 41.25 & 40.82\\
MMOA-RAG-8B & 30.02 & 26.02 & 34.75 & 35.56 & 31.89 & 38.58 & 16.47 & 13.36 & 18.82 & 38.85 & 34.75 & 48.59 & 34.47 & 31.76 & 42.01 & 58.19 \\
AceSearcher-8B & 42.91 & 37.19 & \textbf{68.07} & 50.83 & 45.59 & 52.75 & \underline{25.24} & 19.10 & 26.62 & -- & -- & -- & 47.21 & 45.12 & 51.91 & 59.37 \\
\midrule
\multicolumn{17}{l}{\includegraphics[height=1em]{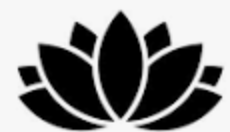} \textit{Ours} } \\
\midrule
\rowcolor{cyan!10} HERA (\qwen) & \textbf{55.38} & \textbf{52.50} & \underline{63.03} & \textbf{60.02} & \textbf{59.50} & \textbf{64.77} & \textbf{27.19} & \textbf{26.57} & \textbf{35.82} & \textbf{54.43} & \textbf{48.74} & \textbf{67.81} & \underline{49.01} & \underline{46.46} & \underline{60.53} & \textbf{67.35} \\
\rowcolor{cyan!10} HERA (\llama) & 53.28 & 48.01 & 57.91 & \underline{52.62} & 46.70 & 55.42 & 25.04 & \underline{24.46} & \underline{31.07} & 48.53 & 43.46 & 60.72 & 46.60 & 45.22 & 56.31 & 60.31 \\
\bottomrule
\end{tabular}}
\end{table}

\vspace{-8pt}

Table~\ref{tab:main_exp_res} compares \hero with a range of baselines (Direct inference and CoT results are reported in Appendix~\ref{app: ablte_llama}).  key observations are: \textbf{(i) Consistent gains over SOTA iterative and agentic RAG baselines}. \hero achieves an average of 38.69\% gains on all datasets compared against recent SOTA, demonstrating the effectiveness of experience-guided orchestration for interleaved reasoning and retrieval. Best AmbigQA performance highlights \hero’s robustness to queries with multiple plausible interpretations. \textbf{(ii) Robust OOD performance.} \hero ranks second on Bamboogle (F1 $-5.4\%$ vs. ExSearch) and first on HoVer ($+64.95\%$ vs. CORAG), reflecting the transferability of the accumulated experience library and evolving agent behaviors. In contrast, most prior approaches relying on heuristic planning or fixed iterative retrieval generalize poorly beyond in-distribution data. \textbf{(iii) Efficiency as gradient-free optimization}. \hero surpasses RL-fine-tuned multi-agent RAG methods, such as MMAO-RAG, showing that semantic group-wise advantages enable stable, consistent improvements across tasks and distribution shifts without costly model updates. \textbf{(iv) \hero-Qwen consistently outperforms \hero-Llama across all datasets}, particularly on cross-document reasoning and complex disambiguation tasks. While \hero-Llama remains competitive in-domain, its performance drops on OOD tasks, indicating limited reasoning and generalization capacity for ambiguous or sparse queries.

\vspace{-8pt}
\subsection{Ablation Study}
\label{sec:ablation}
\vspace{-5pt}
\begin{figure}[ht]
\vspace{-7pt}
        \centering
        \includegraphics[width=1\linewidth]{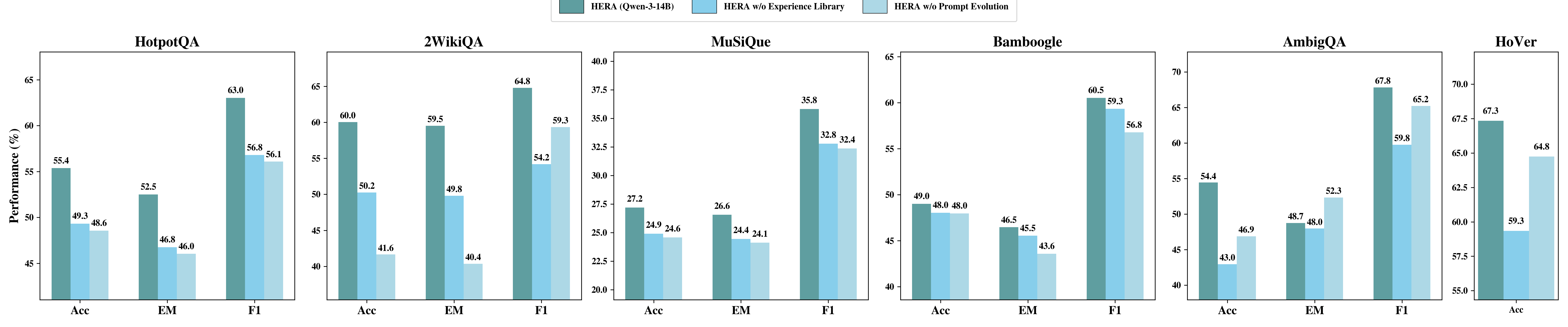}
        \vspace{-8pt}
        \caption{Ablation Studies of \hero with \qwen as the backbone. }
        \label{fig:ablat_qwen}
        \vspace{-15pt}
    \end{figure}
 
Figure~\ref{fig:ablat_qwen} presents the ablation results of \hero on the test set using \qwen as the backbone (\llama results are provided in Appendix~\ref{app: ablte_llama}). Removing the experience library yields consistent but moderate performance declines across datasets (typically $\downarrow$ 6\% to 15\% relative accuracy), highlighting its role in improving knowledge guiding and cross-instance generalization. In contrast, omitting prompt evolution leads to substantially larger and more variable declines on multi-hop QA (up to $\downarrow$ 30\% relative on 2WikiQA for both backbones), where precise decomposition and trajectory control are essential. On ambiguity-intensive datasets AmbigQA, prompt evolution introduces a recall-oriented bias: improving F1 while occasionally reducing EM, whereas the experience library promotes precision through grounded outputs, revealing a systematic EM–F1 trade-off. These patterns are consistent across backbones, indicating that the functional contributions of each component are largely model-agnostic. Importantly, the full \hero framework consistently outperforms both ablations by a non-additive margin, indicating a positive interaction effect: improved reasoning behaviors from prompt evolution enhance the quality of stored experiences, which in turn reinforce subsequent reasoning through stronger grounding. Overall, the results demonstrate that prompt evolution and the experience library target complementary failure modes: procedural correctness versus epistemic reliability, and that their synergy is critical for robust multi-hop reasoning.
\vspace{-8pt}
\subsection{Performance-Efficiency Trade-off}
\label{sec:token_usage}
We examine the potential trade-off between performance and data efficiency by analyzing token usage patterns throughout learning. Our analysis emphasizes two key dimensions: (i) the dynamics of token consumption as learning progresses, and (ii) the interplay between token efficiency and task performance..

\label{sec:token_train}

\setlength{\intextsep}{-1pt}
\begin{wrapfigure}{r}{0.3\textwidth} % l = left, r = right
    \centering
    \includegraphics[width=\linewidth, height=3.6cm, keepaspectratio]{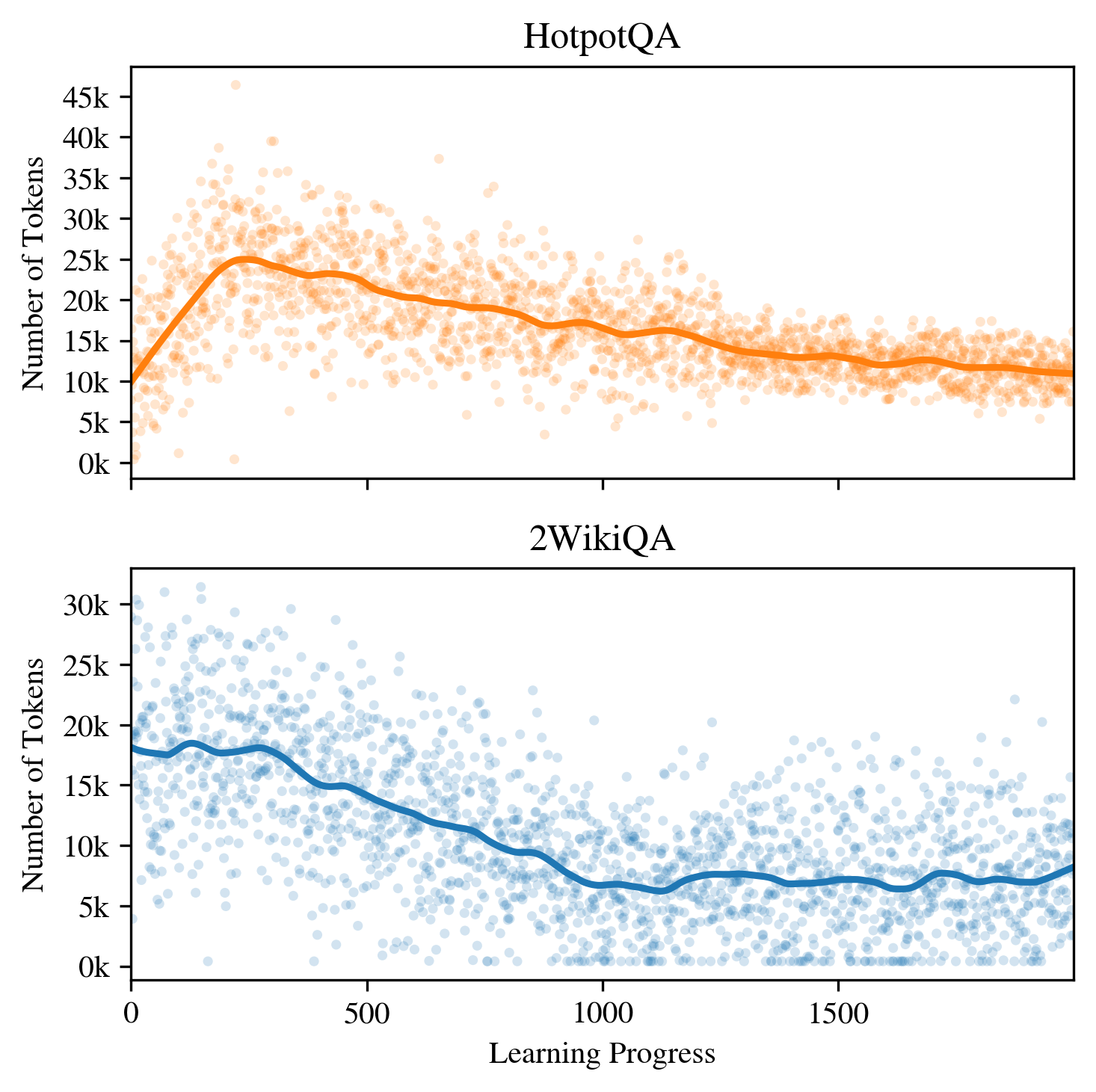}
    \caption{Token usage}
    \label{fig:lowess1}
\end{wrapfigure}
\vspace{-8pt}
\subsubsection{Dynamics of Token Consumption}
\vspace{-5pt}
\label{sec:token_train}
We visualize the token consumption throughout the learning process by Locally Weighted Scatterplot Smoothing (LOWESS) \citep{cleveland1981lowess, dang2025multi}. Figure~\ref{fig:lowess1} illustrates token usage trends on HotpotQA and 2WikiQA as representative tasks. \textbf{(i) Exploration–Exploitation Transition}. On HotpotQA, token consumption follows a non-monotonic trajectory. Early spikes reflect an exploratory phase in which the orchestrator activates multiple agents, constructs extended reasoning traces, and performs redundant steps to optimize performance. As experience accumulates, token usage declines monotonically, marking a transition to exploitation: low-efficiency agents are pruned, and reasoning chains are shortened. A subsequent plateau indicates convergence toward an efficient token budget. In contrast, 2WikiQA exhibits no initial exploratory peak, suggesting that transferable priors from previously accumulated reasoning experience support efficient agent orchestration with minimal trial-and-error. \textbf{(ii) Functional Contributions. } Token usage dynamics are tightly coupled with \hero’s experience library and prompt evolution. The library accumulates high-utility and generalizable experience, while evolving prompts enables agents to elevate their performance. Together, they drive emergent computational frugality, enabling high reasoning performance without explicit token-based penalties.
\vspace{-8pt}
\subsubsection{Performance vs.Token Usage}
\label{sec:performa_vs_token}
\begin{figure}[h]
        \centering
        \includegraphics[width=1\linewidth]{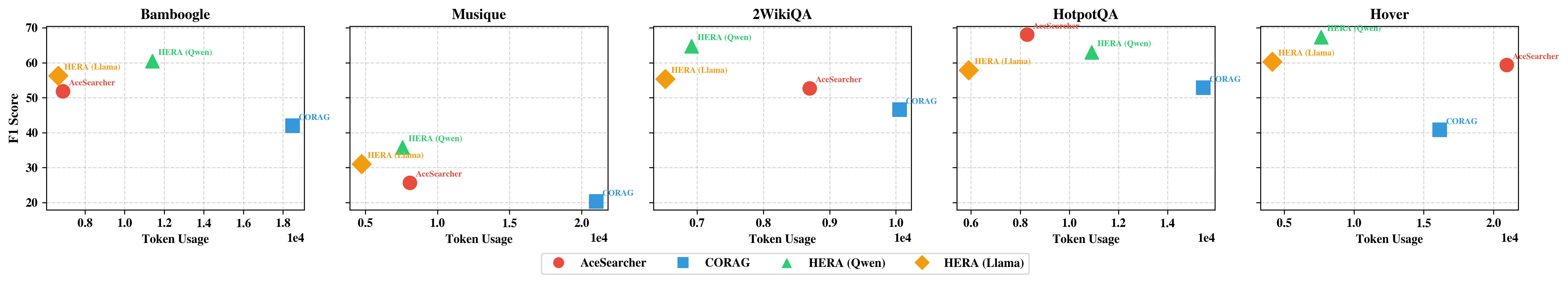}
        \vspace{-12pt}
        \caption{Comparison of Performance–Token Efficiency Trade-off with Selected Baselines.}
        \label{fig:f1_tokens}

    \end{figure}
\vspace{10pt}
Figure~\ref{fig:f1_tokens} compares multi-hop QA performance against token consumption. Key observations include: \textbf{(i) Superior Performance–Efficiency Trade-off}. Across all datasets, \hero consistently dominates the performance–efficiency frontier, achieving higher F1 scores with controlled token budgets. This demonstrates that gains arise from efficient reasoning trajectories instead of brute-force context scaling.
\textbf{(ii) Backbone-Specific Patterns.)} \hero with \qwen achieves the highest absolute performance, whereas \hero with \llama consistently minimizes token consumption ($\sim$4.7k–6.6k across most datasets) while preserving competitive F1 scores, indicating favorable computational efficiency. \textbf{(iii) Consistent performance and efficiency gains over baselines}.  CORAG uses the largest number of tokens across most datasets (up to $20k+$) without converting these costs into sustainable reasoning improvements or effective knowledge reuse, resulting in suboptimal F1 (\eg, 20.30 on MusiQue and 40.82 on HoVer). AceSearcher occupies an intermediate position, balancing moderate token usage and performance. 
\vspace{-8pt}
\subsection{Topology Evolution}
\vspace{-5pt}
\label{sec:topology_analysis}
To elucidate the emergence of coordinated multi-agent behavior, we systematically analyze the evolution of interaction topologies over the course of learning. For this purpose, we introduce topology-based metrics to quantify structural changes in agent collaboration.

\vspace{-5pt}
\subsubsection{Transition Entropy}
\label{sec:transition_entropy}
\begin{wrapfigure}{r}{0.32\textwidth}
    \vspace{-20pt} 
    \centering
    \includegraphics[width=\linewidth, height=3.3cm, keepaspectratio]{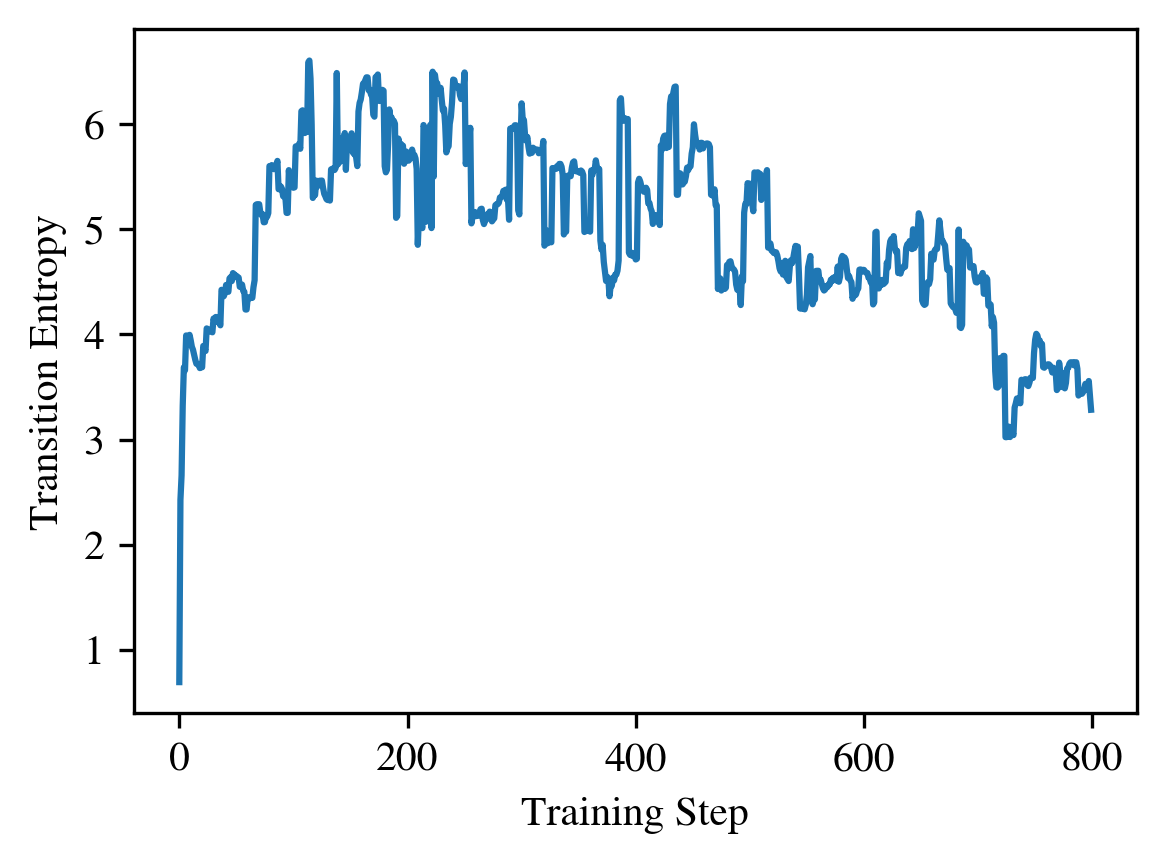}
    \vspace{-20pt} 
    \captionof{figure}{Transition entropy}
    \vspace{-20pt} 
        \label{fig:tran_entr}
        
\end{wrapfigure}
Transition Entropy quantifies the uncertainty in agent-to-agent transitions, capturing the policy-level dynamics. Let $Prob.(\mathcal{N}_{j}|\mathcal{N}_{i})$ denote the empirical transition probability from agent $\mathcal{N}_{i}$ to $\mathcal{N}_{j}$, the Transition Entropy is defined as:

\begin{flushleft}
\(
H_{trans} = -\sum_{i,j} Prob.(\mathcal{N}_{i} \to \mathcal{N}_{j})\,\log Prob.(\mathcal{N}_{i} \to \mathcal{N}_{j})
\)
\end{flushleft}
We compute $H_{trans}$ using a sliding-window over learning. Beyond the exploration–exploitation dynamics, Figure~\ref{fig:tran_entr} shows that after an initial rise, entropy stabilizes at an intermediate level, reflecting the emergence of structured yet flexible agent interactions. This plateau indicates that \hero consolidates high-value transition sequences while maintaining controlled exploration, enabling the system to integrate novel reasoning strategies without disrupting established coordination. The pattern highlights the complementary roles of the experience library, which reinforces effective multi-agent pathways, and prompt evolution, which adaptively refines agent behavior for functional alignment with task objectives. Notably, the plateau at intermediate entropy levels indicates controlled exploration even during exploitation, preventing premature convergence to suboptimal strategies while allowing the integration of novel cooperation pathways.

\vspace{-5pt}
\subsubsection{Graph-based Metrics}
\begin{wrapfigure}{r}{0.32\textwidth}

    \centering
    \includegraphics[width=\linewidth, height=3.3cm, keepaspectratio]{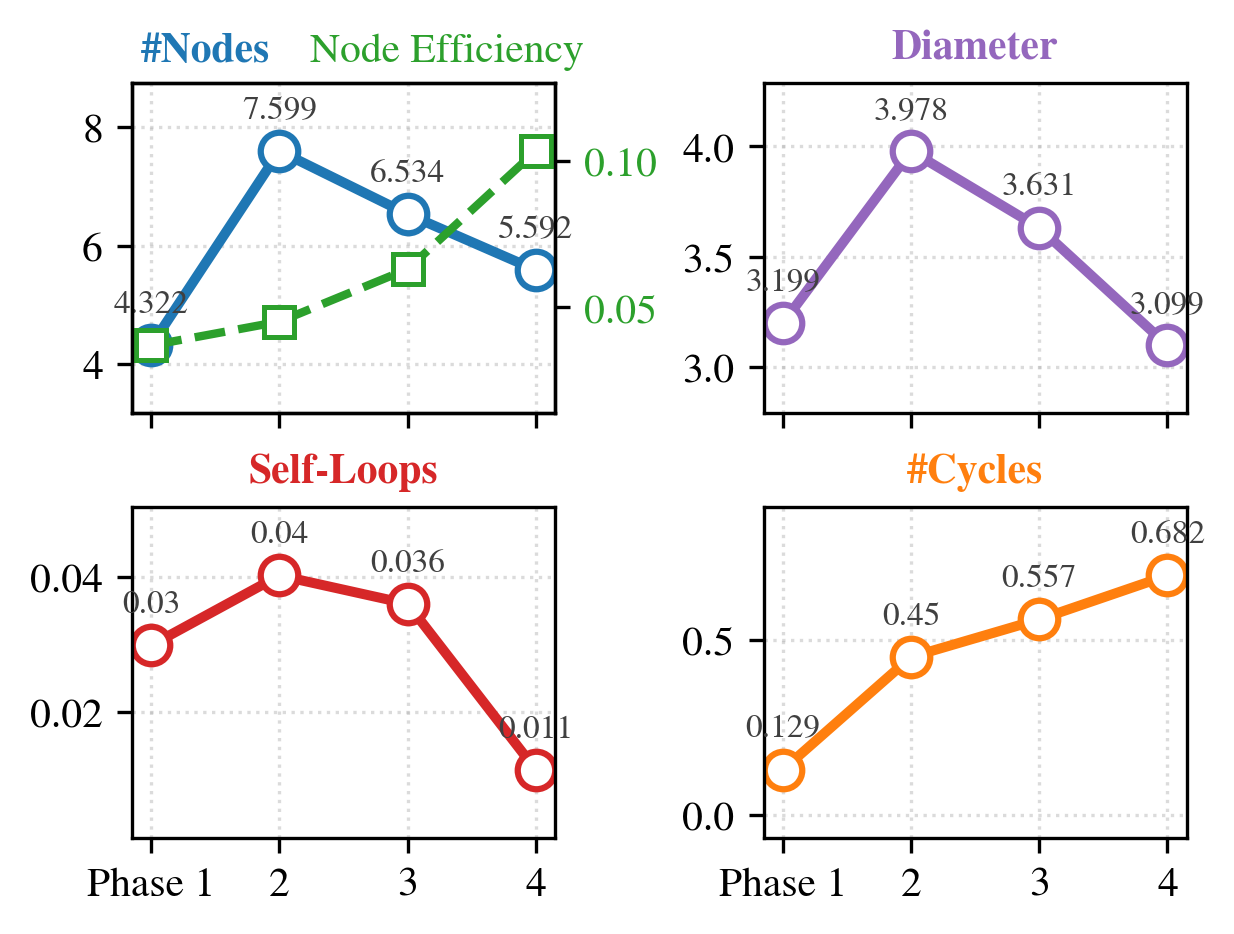}
    \captionof{figure}{Graph metrics}

    \label{fig:graph_metrics}
\end{wrapfigure} 
\vspace{-5pt}
To systematically characterize the topology evolution of \hero, we model each trajectory as a graph $\mathcal{G}_{\tau}=(V, E)$ where nodes represent agents and edges denote interactions. We quantify structural and functional properties via defining graph-theoretic metrics:
\textbf{(a) Number of agents $|V|$}: the total distinct agents involved in a trajectory, reflecting the breadth of collaboration.
\textbf{(b) Node efficiency}: per-agent contribution, formally by $\frac{\text{F1}(\tau)}{|V|}$, capturing per-agent utility.
\textbf{(c) Self-Loops}: the frequency of self-transitions, reflecting potential redundancy or degenerate reasoning.
\textbf{(d) Number of cycles}: count of closed-loop paths capturing iterative reasoning patterns.
\textbf{(e) Diameter}: longest shortest-path between any two nodes, representing maximum reasoning depth. 

We segment learning into four phases - initial, exploration, refinement, and optimization, and analyze normalized metrics per question type. As shown in Fig.~\ref{fig:graph_metrics}, Early trajectories are narrow and shallow, with low agent counts, minimal cycles, and limited diameter, reflecting sparse coordination. During exploration, the system dynamically recruits more agents, increasing diameter and cycle formation, capturing iterative reasoning and intermediate verification. By the refinement phase, trajectories contract: redundant agents and self-loops are pruned, diameter decreases, and node efficiency rises. This selective retention is guided by the experience library, which reinforces high-value interaction topologies, and by prompt evolution, which refines agent behaviors and intermediate outputs. The optimized phase exhibits compact chains with minimal nodes, high per-agent utility, and strategically preserved cycles, representing a balance of efficiency and robustness.

Overall, the evolving topologies reveal that \hero’s dual mechanisms - structured experience retrieval and adaptive prompt evolution enable a principled transition from exploratory, diffuse agent interactions to compact, high-utility, and resilient multi-agent system.

\vspace{-10pt}
\section{Related Work}
\vspace{-10pt}
\label{sec:related_work}
\textbf{Retrieval-Augmented Generation (RAG)} boosts LLM performance by leveraging external evidence \citep{izacard2023atlas, lewis2020retrieval, gao2023retrieval, tang2024multihoprag, xiong-etal-2024-benchmarking, fan2024survey}. Early methods embed retrieval within model architectures, permitting direct conditioning on retrieved context \citep{Khandelwal2019GeneralizationTM, liu2024chatqa, izacard2020distilling, wang2023shall, borgeaud2022improving, izacard2023atlas, guu2020retrieval}. Subsequent methods modularize RAG into pre- \citep{chan2024rq, gao-etal-2023-precise, ma-etal-2023-query} and post-retrieval components \citep{abdallah2025rankify, xu2023recomp, li2025oreo} to enhance retrieval and reasoning, and synergize retrieval with structured reasoning for multi-step tasks \citep{jeong-etal-2024-adaptive, lee-etal-2024-planrag, asai2024self, shao-etal-2023-enhancing}. Emerging paradigms such as DeepResearcher \cite{zheng-etal-2025-deepresearcher, li2025webthinker, huang2025deep} extend RAG into agentic \citep{xie2025marsha, fang2025orion, chen2025mao, chen2025improving} workflows, embedding retrieval within iterative reasoning \citep{chen2025learning, shi2025iterative, li2025search, jin2025search, song2025r1} and tool use \citep{song2025r1, sun2025simpledeepsearcher}, to tackle complex problems.

\vspace{-4pt}
\textbf{Multi-agent Orchestration}  
Prior works \citep{chen2023agentverse, liu2024dynamic, qian2024scaling, khan2024debating, liang2024encouraging, wang2024rethinking, iqbal2022alma} often frame orchestration as a sequential pipeline optimized via reinforcement learning \citep{nie2025weak, wang2026agentconductor}. Dynamic methods adapt agent interactions at inference based on task features or intermediate outcomes \citep{yu2026adaptorch, yuan2025automated, zhang2026verified, chang2025sagallm}, but incur high coordination overhead and rely on accurate intermediate evaluations. Alternatively, some approaches treat orchestration as a global decision problem, learning a unified policy over the full interaction structure \citep{ke2026mas, Zhang2025AgentOrchestraOM}. Prompting strategies and tool management improve agent coordination \citep{lin2025mao, su2025toolorchestra, Zhang2025AgentOrchestraOH, dang2025multi, Du2024MultiAgentCV} but current approaches are often static \citep{zhou2026orch}, weakly adaptive, or limited by computational and scalability constraints \citep{choi2024malade}. 

\vspace{-4pt}
\textbf{Prompt Optimization and Evolution} 
Automatic Prompt Optimization treats prompts as a black-box search over seeds, feedback, candidate generation and selection \citep{ramnath2025systematic}. 
Programmatic frameworks \citep{khattab2024dspy} compile LM pipelines and optimize prompts offline using user metrics, while textual backpropagation \citep{yuksekgonul2025optimizing, hu2024self, ma2025agentic} utilizes natural-language feedback to refine collaboration protocols and behavioral instructions. Recent work jointly optimizes agentic structures and prompts \citep{tian-etal-2025-agentinit, murthy2025promptomatix, spiess2025autopdl}, using search-based strategies \citep{zhang2024aflow}, genetic algorithms \citep{agrawal2025gepa} and treats prompt optimization as experience-driven adaptation \citep{yao2022react, shinn2023reflexion, zhao2024expel, pei2025scope, xie2026memo}.

\section{Conclusion}

\label{sec:cond}
We present \hero, a hierarchical framework that jointly evolves multi-agent orchestration through an experience library and role-aware agent prompts. The experience library consolidates transferable orchestration strategies, while prompt evolution enables role-specific refinement of agent behaviors along operational and behavioral principles. \hero achieves superior  performance on multi-hop and ambiguity-intensive benchmarks without relying on gradient-based training. Topological analyses reveal emergent self-organization, where sparse exploration gives rise to compact, high-utility multi-agent networks, highlighting both efficient coordination and robustness. Future work may investigate meta-optimization of agent roles, adaptive multi-turn orchestration, and resource-aware autonomous evolution, paving the way for hierarchical role structures, conditional agent sequencing, and efficient reasoning under token or latency constraints.

\section*{Acknowledgments}
Use unnumbered first level headings for the acknowledgments. All
acknowledgments, including those to funding agencies, go at the end of the paper.

\section*{Ethics Statement}
Authors can add an optional ethics statement to the paper. 
For papers that touch on ethical issues, this section will be evaluated as part of the review process. The ethics statement should come at the end of the paper. It does not count toward the page limit, but should not be more than 1 page.

\bibliography{colm2026_conference}
\bibliographystyle{colm2026_conference}

\newpage

\appendix
\section{Theoretical Analysis}
\label{app:theory}
\subsection{Formal Setup}
Let $q \in \mathcal{Q}$ denote an input query, $\mathcal{E}$ the experience library, $\mathcal{N}$ the agent pool, $\Gamma$ an agent topology specifying the agent sequence, orders, and dependencies, and $\tau$ an execution trajectory. The orchestrator induces a policy to obtain an interaction topology:
\[
\Gamma \sim \pi_{\theta}(\cdot \mid q, \mathcal{E}, \mathcal{N}).
\]

\subsection{Policy Iteration}
\hero admits a rigorous interpretation as gradient-free policy iteration over the structured space $\Gamma$. Each iteration $t$ comprises two steps:

\paragraph{\textbf{Policy Evaluation.}}
A group of rollouts/topologies is sampled from the current policy $\pi_{\mathcal{O}}$ and executed to obtain trajectories and rewards. This constitutes an empirical Monte Carlo estimate of the value function under the current policy. 

\paragraph{\textbf{Policy Improvement.}}
Group-relative semantic advantages are obtained by reflecting on and comparing high-reward trajectories with failed trajectories. These insights are distilled and incorporated into the experience library $\mathcal{E}$:
\[
\mathcal{E}^(t+1) \gets \text{Distill}\left(\{(\Gamma_{i}, \tau_i) : R_i > \bar{R}\}\right)
\] where $\bar{R}$ is the semantic baseline. 

This updated experience library  modifies the conditioning context, which in turn reshapes the induced distribution: 
\[
\pi_{\pi_\mathcal{O}}\bigl(\Gamma \mid q, \mathcal{E}^{(t+1)}, \mathcal{N}\bigr)
\] without updating $\pi_\mathcal{O}$.

\subsection{EM Interpretation}
\hero exhibits an iterative structure analogous to Expectation-Maximization. 

The agent topology $\Gamma$ as a latent variable, the query $q$ as observed input, and the reward $R(\tau)$ as a proxy for an implicit likelihood. The iterative procedure then resembles:

\medskip
\noindent\textbf{1. E-step (Implicit)}: Sampling 
\[
\Gamma_{i} \sim {\pi_\mathcal{O}}(\cdot \mid q, \mathcal{E}^{(t)}, \mathcal{N})
\] 
and selecting high-reward topologies induces an approximate posterior over effective reasoning programs:
\[
\mathcal{P}^{(t)}(\Gamma) \approx 
\frac{\pi_{\mathcal{O}}(\cdot \mid q, \mathcal{E}^{(t)}, \mathcal{N}) \cdot \mathbf{}[R(\tau_i) > \bar{R}]}%
{\sum_{\Gamma'} \pi_{\theta_o}(\Gamma' \mid q, \mathcal{E}^{(t)}, \mathcal{N})) \cdot \mathbf{1}[R(\tau_{\Gamma'}) > \bar{R}]}
\]
This represents a reward-filtered empirical distribution instead of a properly normalized probabilistic posterior.

\noindent\textbf{M-step(Implicit)}:instead of maximizing
\[
\mathbb{E}_{\mathcal{P}^{(t)}(S)} \Bigl[ \log p\bigl(y \mid q, \Gamma; \pi_{\mathcal{o}} \bigr) \Bigr]
\]
over parameters $\pi_{\mathcal{O}}$, \hero performs a non-parametric update:
\[
\mathcal{E}^{(t+1)} \leftarrow 
\arg\max_{\mathcal{E}} \, 
\mathbb{E}_{\mathcal{P}^{(t)}(\Gamma)} \bigl[\log \pi_{\theta_\mathcal{O}}(\Gamma \mid q, \mathcal{E}, \mathcal{N})\bigr]
\]
which implicitly increases the probability mass on high-performing structures by enriching the conditioning context (experience library).
$R(\tau)$ here serves as a heuristic surrogate rather than a proper probabilistic objective.

\subsection{Energy-Based Reweighting and Implicit KL Regularisation}

The frozen base model $\pi_{\mathcal{O}}$ provides implicit regularisation. As policy updates are mediated through the contextual augmentation $\mathcal{E}$ instead of modifying parameters, the resulting policy can be viewed as an energy-based reweighting of the base distribution:

\[
\pi_{\text{new}}(\Gamma \mid q) \propto \pi_{\text{base}}(\Gamma \mid q) \cdot \exp\bigl(f_{\mathcal{E}}(\Gamma, q)\bigr),
\]

where $f_{\mathcal{E}}(\Gamma, q)$ denotes the log-ratio bias introduced by the experience library. This formulation is structurally equivalent to the optimal solution of the KL-constrained policy optimisation problem:

\[
\max_{\pi} \, \mathbb{E}_{\pi}[f_{\mathcal{E}}(\Gamma, q)] - \beta \cdot D_{\mathrm{KL}}\bigl(\pi \,\|\, \pi_{\text{base}}\bigr),
\]

whose closed-form solution is precisely

\[
\pi^*(\Gamma \mid q) \propto \pi_{\text{base}}(S \mid q) \cdot \exp\Bigl(\frac{f_{\mathcal{E}}(\Gamma, q)}{\beta}\Bigr).
\]

The frozen base model therefore serves as an implicit KL anchor, preventing policy collapse and ensuring that the updated distribution stays within the support of the original model. This stability property is absent in unconstrained policy gradient methods, where aggressive updates can destabilise the base distribution.
\newpage
\section{Prompts used for \hero}
\label{app:promptt}
\begin{tcolorbox}[
    colback=promptbg!100!white,
    sharp corners=south, 
    boxrule=0.25mm, 
    fonttitle=\bfseries, 
    title={\textbf{Orchestrator: topology sampling}}, 
    width=\textwidth, 
    enhanced,
    drop shadow
    ]

You are an orchestrator managing a team of specialized agents for multi-hop 
question answering. Your goal is to design an effective multi-agent execution topology 
to answer the given query correctly. \\[2mm]
\textbf{You have access to the following agents:} \\
\{agent\_descriptions\}

\vspace{0.5em}
\textbf{You have retrieved the following relevant experiences from past executions:} \\
\{retrieved\_experiences\}

\vspace{0.5em}
\textbf{Each experience entry has the format:}
\begin{itemize}
    \item Query Type: \{query\_type\}
    \item Insight: \{insight\}
    \item Utility score: \{utility\}
\end{itemize}

\vspace{0.5em}
\textbf{Given the query below, design a coordination topology by:}
\begin{enumerate}
    \item Selecting the subset of agents best suited to this query type.
    \item Specifying their execution order (sequential or parallel where appropriate).
    \item Defining dependency relationships (which agent's output feeds into which).
\end{enumerate}

\vspace{0.5em}
\textbf{Query:} \{query\}

\vspace{0.5em}
\textbf{Respond in the following JSON format:}
\begin{verbatim}
{
  "query_profile": "<one-sentence characterization of query type>",
  "selected_agents": ["<agent_name>", ...],
  "execution_order": [
    {"step": 1, "agent": "<agent_name>", "depends_on": [],
    
    "mode": "sequential|parallel"},
    ...
  ]
}
\end{verbatim}
\end{tcolorbox}

\begin{tcolorbox}[colback=promptbg!100!white,title=Orchestrator: Semantic Advantage Extraction]
\small
You are an orchestrator evaluating a group of multi-agent execution trajectories
for the same query. Your goal is to identify why some trajectories succeeded and 
others failed, and extract generalizable insights to guide future orchestration.

\vspace{0.5em}
\textbf{Query:} \{query\} \\
\textbf{Query type:} \{query\_type\}

\vspace{0.5em}
The following \{G\} trajectories were executed, ranked by task performance (F1) 
and then by efficiency (total tokens consumed):

\vspace{0.5em}
\{trajectory\_group\}

\vspace{0.5em}
\textbf{Each trajectory entry includes:}
\begin{itemize}
    \item Topology used (agents selected, execution order)
    \item Per-agent actions, inputs and outputs
    \item Final answer produced
    \item F1 score: \{f1\}
    \item Total tokens consumed: \{tokens\}
\end{itemize}

\vspace{0.5em}
\textbf{Perform a comparative analysis across the group:}
\begin{enumerate}
    \item Identify what the successful trajectories did differently from the failed ones.
    \item Identify which agents, orderings, or dependencies contributed to success or failure.
    \item Identify recurring failure patterns (e.g., incorrect agent selection, redundant steps, missing retrieval before reasoning).
    \item Distill findings into concise, actionable insights applicable to future queries of the same type.
\end{enumerate}

\vspace{0.5em}
\textbf{Respond in the following JSON format:}
\begin{verbatim}
{
  "success_factors": ["<factor>", ...],
  "failure_modes": ["<failure_mode>", ...],
  "insights": [
    {
      "query_type": "<type this insight applies to>",
      "insight": "<actionable natural language insight>"
    },
    ...
  ]
}
\end{verbatim}
\end{tcolorbox}

\begin{tcolorbox}[
    colback=promptbg!100!white,
    sharp corners=south, 
    boxrule=0.25mm, 
    fonttitle=\bfseries, 
    title={\textbf{Experience Library Operations)}}, 
    width=\textwidth, 
    enhanced,
    drop shadow
    ]
You are managing an experience library for a multi-agent question answering system. 
The library stores insights as structured entries of the form:
\[
(\text{query\_profile } c,\ \text{insight } z,\ \text{utility } u)
\]
where utility $u$ reflects how often insight $z$ has led to successful agent orchestration.

\vspace{0.5em}
\textbf{Current experience library:} \\
\{current\_library\}

\vspace{0.5em}
\textbf{New insights extracted from the latest execution group:} \\
\{new\_insights\}

\vspace{0.5em}
\textbf{For each new insight, decide the appropriate consolidation operation:}
\begin{itemize}
    \item \textbf{ADD}: Insert a distinct insight not covered by existing entries.
    \item \textbf{MERGE}: Combine semantically similar or complementary insights into a single, more complete entry.
    \item \textbf{PRUNE}: Remove a lower-utility entry that conflicts with a higher-utility one.
    \item \textbf{KEEP}: Retain the current library without modification.
\end{itemize}

\vspace{0.5em}
\textbf{Guidelines:}
\begin{itemize}
    \item Prefer \textbf{MERGE} over \textbf{ADD} when insights share the same query type and recommend compatible strategies.
    \item Prefer \textbf{PRUNE} when entries provide contradictory guidance for the same query type and differ substantially in utility.
    \item Avoid unbounded growth—consolidate aggressively to maintain generalizability.
\end{itemize}

\vspace{0.5em}
\textbf{Respond in the following JSON format:}
\begin{verbatim}
{
  "operations": [
    {
      "operation": "ADD|MERGE|PRUNE|KEEP",
      "new_insight": "<text of new insight>",
      "target_entry_ids": ["<id of existing entry if MERGE or PRUNE>"],
      "merged_insight": "<combined insight text if MERGE, else null>",
      "rationale": "<one sentence explaining the decision>"
    },
    ...
  ]
}
\end{verbatim}
\end{tcolorbox}

\begin{tcolorbox}[
    colback=promptbg!100!white,
    sharp corners=south, 
    boxrule=0.25mm, 
    fonttitle=\bfseries, 
    title={\textbf{Prompt for Agent Prompt Integration}}, 
    width=\textwidth, 
    enhanced,
    drop shadow
    ]

\textbf{System Prompt:} \\
You are responsible for managing and evolving the prompt of a single agent under strict length and structural constraints. Your task is to refine, optimize, and consolidate the agent’s prompt to ensure clarity, consistency, and adherence to operational and behavioral requirements. \\[2mm]

\textbf{User Prompt:} \\

\textbf{Current Agent Prompt:} \\
\begin{verbatim}
{CURRENT_PROMPT}
\end{verbatim}

\textbf{Proposed Operational Rules:} \\
\begin{verbatim}
{operational_rules}
\end{verbatim}

\textbf{Proposed Behavioral Principles:} \\
\begin{verbatim}
{behavioral_principles}
\end{verbatim}

\textbf{Constraints:} 
\begin{enumerate}
    \item Limit the number of tactical rules to a maximum of \textbf{{K}}.  
    \item All instructions must be internally consistent — no contradictions.
    \item Preserve the agent's core role definition and tool usage instructions
    \item Remove redundant, contradictory, or ambiguous instructions.  
    \item Preserve essential operational and behavioral requirements.  
    \item Ensure the updated prompt is concise, coherent, and actionable.  
\end{enumerate}

\textbf{Task:} \\
Produce a \textbf{prompt diff} that clearly indicates the modifications required to integrate the proposed rules and principles into the current agent prompt while satisfying the constraints above. Highlight additions, deletions, and replacements in a structured format.

\end{tcolorbox}

\begin{tcolorbox}[
    colback=promptbg!100!white,
    sharp corners=south, 
    boxrule=0.25mm, 
    fonttitle=\bfseries, 
    title={\textbf{RoPE: contrastive analysis}}, 
    width=\textwidth, 
    enhanced,
    drop shadow
    ]

You are analyzing the results of prompt variant re-executions for a failing agent to extract concrete prompt improvements.

\textbf{Agent name:} \{agent\_name\} \\
\textbf{Agent role:} \{agent\_role\} \\
\textbf{Original prompt:} \{original\_prompt\}

\textbf{Variant execution results:} \\
\{variant\_results\}

Each result entry contains:
\begin{itemize}
  \item \textbf{Variant prompt:} \{variant\_prompt\}
  \item \textbf{Re-executed trajectory:} \{trajectory\}
  \item \textbf{F1 score:} \{f1\}
  \item \textbf{Tokens used:} \{tokens\}
\end{itemize}

Perform a contrastive analysis between successful and failed variants:

\begin{enumerate}
  \item \textbf{Operational rules} ($\Delta \rho_i^{op}$): Extract short-term corrective behaviors — specific, concrete instructions that directly address the observed failure pattern. These should be actionable in the agent's very next execution. \\

  \item \textbf{Behavioral principles} ($\Delta \rho_i^{bp}$): Extract long-term strategic generalizations — higher-level guidance distilled from patterns across multiple trajectories. These shape the agent's overall approach rather than fixing a single failure. \\

\end{enumerate}
Respond in the following JSON format:
\begin{verbatim}
{
  "operational_rules": [
    {
      "rule": "<concrete instruction>",
      "derived_from": "<which variant comparison motivated this rule>"
    },
    ...
  ],
  "behavioral_principles": [
    {
      "principle": "<strategic guidance>",
      "derived_from": "<pattern observed across which variants>"
    },
    ...
  ],
  "updated_prompt": "<full revised prompt integrating all rules and principles \\ 
  into the original, with redundant instructions pruned>"
}
\end{verbatim}
\end{tcolorbox}

\section{Datasets}
\label{app:dataset}
We use the following datasets in our experiments: 
1. 2WikiMultihopQA \citep{ho-etal-2020-constructing}: a multi-hop QA dataset that constructs questions using both Wikipedia text and structured Wikidata, explicitly providing annotated reasoning paths (evidence chains) to ensure and evaluate true multi-step reasoning across multiple documents.

2. HotPotQA \citep{yang2018hotpotqa}: a multi-hop QA dataset built from Wikipedia that requires models to reason over multiple documents while providing sentence-level supporting facts to enable explainable answer prediction

3. Bamboogle \citep{press-etal-2023-measuring}: a small, manually constructed QA dataset of 125 challenging real-world multi-hop questions designed such that answers cannot be directly retrieved from search engines, requiring compositional reasoning across multiple pieces of evidence.

4. MusiQue \citep{trivedi2021musique}: a multi-hop QA dataset constructed by composing connected single-hop questions into 2–4 hop reasoning chains, explicitly designed to enforce genuine multi-step reasoning and reduce shortcut-based answering.

5. HoVer \citep{jiang2020hover}: a multi-hop fact verification dataset built from Wikipedia where models must retrieve evidence across 2–4 documents and determine whether a claim is supported or not, emphasizing complex many-hop reasoning and evidence extraction. 

6. Ambig\_QA \citep{min-etal-2020-ambigqa}: an open-domain QA dataset derived from NQ-open that focuses on ambiguous questions, requiring models to generate all plausible answers along with corresponding disambiguated question rewrites to explicitly resolve ambiguity.

\vspace{10pt}
\begin{table}[h]
\centering
\caption{Statistics of datasets.}
\label{tab:dataset_stats}
\begin{tabularx}{\columnwidth}{lccccX}
\toprule
\textbf{Dataset} & \textbf{Train} & \textbf{Val} & \textbf{Test} & \textbf{Total} & \textbf{Reasoning} \\
\midrule
2WikiQA & 154K & 16K & 22K & 192K & 2-hop (cross-doc) \\
HotpotQA        & 90K  & 7.4K & 7.4K & 105K & 2-hop (bridge/comp) \\
MuSiQue         & 19K  & 2.4K & 2.4K & 24K  & 2--4 hop (compositional) \\
Bamboogle       & --   & --   & 125  & 125  & Multi-hop (adversarial) \\
HoVer           & 13K  & 1.8K & 2.7K & 18K  & 2--4 hop (verification) \\
AmbigQA         & 10K  & 2K   & 2K   & 14K  & Ambiguity resolution \\
\bottomrule
\end{tabularx}
\end{table}

\section{Construction of Training Sets}
\label{app:training_set}

\subsection{Question Types}
In this work, we categorize questions into seven types based on the reasoning complexity they require. 
\begin{itemize}
    \item \textit{Bridge multi-hop} questions involve a sequential dependency through an intermediate entity, for example, ``Which university did the author of \textit{The Old Man and the Sea} attend?''.
    \item  \textit{Intersection multi-hop} questions require answers that satisfy multiple independent constraints, such as ``Which scientists won a Nobel Prize and later served as a university president?''.
    \item \textit{Comparison multi-hop} questions compare attributes across entities after retrieval, exemplified by ``Who was born earlier, Marie Curie or Albert Einstein?''.
    \item \textit{Temporal multi-hop} questions necessitate reasoning over time ordering or temporal containment, for instance, ``Who was president of the U.S. when the Berlin Wall fell?''.
    \item \textit{Causal multi-hop} questions involve explaining cause–effect chains across events, as in ``Why did the 2008 financial crisis lead to increased banking regulation?''. 
    \item \textit{Ambiguous} questions are those with multiple plausible interpretations, such as ``When did the Manhattan Project begin and end?''. This taxonomy allows us to analyze reasoning strategies across diverse question structures.

\end{itemize}

\subsection{Distribution}
\begin{figure}[ht]
        \centering
        \includegraphics[width=1\linewidth]{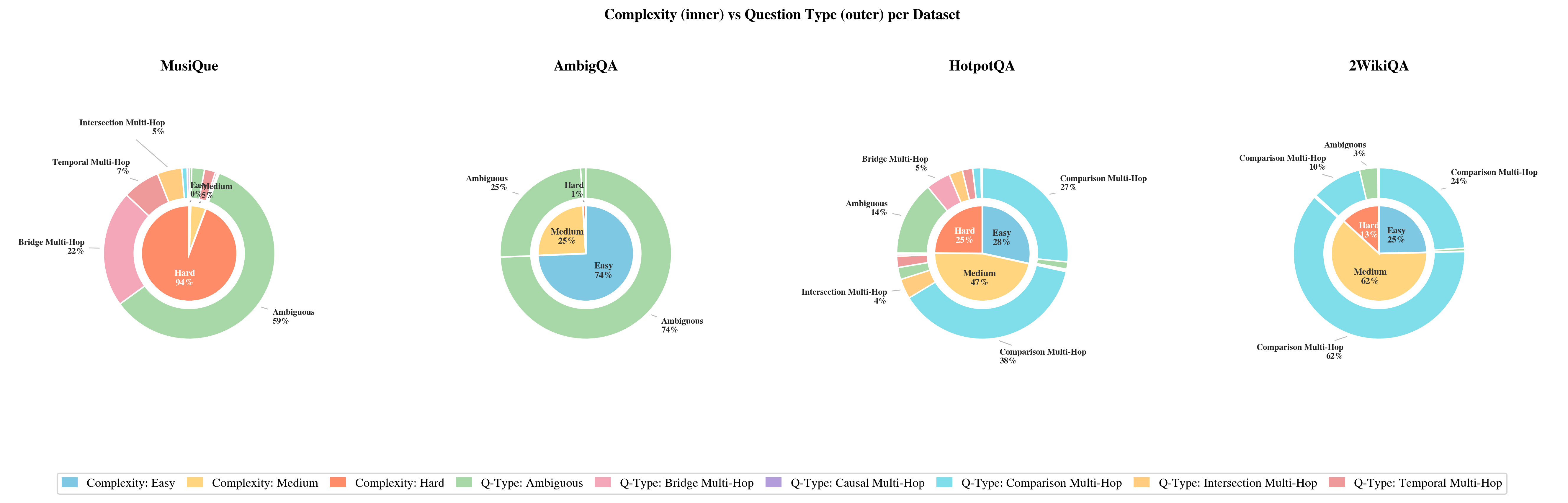}
        \caption{Distribution of Reasoning Types and Complexity of Datasets}
        \label{fig:pie}
\end{figure}

Fig.~\ref{fig:pie} shows the distribution of reasoning types and complexity across the four datasets that are used to construct training data.

\section{Baselines and Implementation Details}
\label{app:baselin_model}
We standardize decoding and prompting across baselines to ensure fair comparison. For LLMs Qwen-3-4B, 8B, and 14B \citep{yang2025qwen3}, Qwen2.5-Instruct-7B \citep{hui2024qwen2}, LLaMA3.1-8B \citep{grattafiori2024llama}, and GPT-4o-mini \citep{hurst2024gpt}, we follow official guidance. We set $temperature = 0.6, top-p = 0.95, top-k = 20, and min-p = 0$.

\section{Additional Experiment Results}

\subsection{Comparison \hero with Direct inference and CoT}
\label{app:additional_res_direct}
\begin{table}[h]
\centering
\scriptsize
\setlength{\tabcolsep}{2pt}
\caption{Additional comparison between \hero and baselines without RAG. }
\label{tab:part1-data-swapped}
\resizebox{\linewidth}{!}{
\begin{tabular}{l|ccc|ccc|ccc|ccc|ccc|c}
\toprule
\textbf{Dataset} 
& \multicolumn{3}{c|}{HotpotQA} 
& \multicolumn{3}{c|}{2WikiQA} 
& \multicolumn{3}{c|}{MusiQue} 
& \multicolumn{3}{c|}{AmbigQA} 
& \multicolumn{3}{c|}{Bamboogle} 
& HoVer \\
\cmidrule(r){2-4} \cmidrule(r){5-7} \cmidrule(r){8-10} 
\cmidrule(r){11-13} \cmidrule(r){14-16} \cmidrule(r){17-17}
\textbf{Metrics} 
& Acc & EM & F1 
& Acc & EM & F1 
& Acc & EM & F1 
& Acc & EM & F1 
& Acc & EM & F1 
& Acc \\
\midrule
\multicolumn{17}{l}{\textbf{\textit{Direct Inference w/o RAG}}} \\
\midrule
Llama3-3.1-8B-instruct & 15.63 & 9.60 & 22.13 & 24.13 & 16.53 & 28.67 & 0.20 & 0.99 & 7.18 & 4.80 & 4.80 & 7.96 & 10.74 & 8.29 & 18.55 & 5.80 \\
Qwen2.5-7B-instruct & 19.30 & 13.17 & 27.97 & 24.43 & 19.70 & 28.88 & 2.61 & 2.40 & 10.72 & 8.70 & 8.00 & 8.80 & 16.73 & 14.04 & 25.65 & 3.08  \\
Qwen3-8B & 17.83 & 11.60 & 26.49 & 24.37 & 18.80 & 28.83 & 2.69 & 2.52 & 10.37 & 11.20 & 9.60 & 18.35 & 14.19 & 11.84 & 23.24 & 5.90 \\
Qwen3-14B & 21.40 & 14.53 & 30.68 & 28.00 & 20.87 & 32.69 & 2.61 & 2.36 & 10.03 & 6.40 & 4.80 & 12.97 & 17.73 & 14.34 & 27.98 & 3.94 \\
\midrule
\multicolumn{17}{l}{\textbf{\textit{CoT w/o RAG}}} \\
\midrule
Llama3.1-8B-instruct & 16.20 & 10.14 & 24.34 & 24.70 & 16.74 & 28.73 & 0.27 & 1.40 & 7.64 & 15.45 & 5.20 & 11.90 & 11.48 & 9.00 & 21.33 & 15.34 \\
Qwen2.5-7B-instruct & 19.93 & 13.73 & 28.24 & 24.37 & 18.00 & 27.94 & 4.34 & 4.14 & 11.47 & 28.00 & 25.60 & 38.16 & 17.88 & 15.23 & 28.14 & 15.93 \\
Qwen3-8B & 21.30 & 15.47 & 29.36 & 25.17 & 17.63 & 29.51 & 3.68 & 3.56 & 11.03 & 29.60 & 23.20 & 39.83 & 16.78 & 14.29 & 26.73 & 20.10 \\
Qwen3-14B & 19.80 & 13.27 & 28.48 & 25.33 & 19.03 & 29.70 & 2.61 & 2.44 & 10.03 & 12.00 & 10.40 & 19.38 & 14.89 & 12.99 & 24.38 & 22.09 \\
\midrule
\multicolumn{17}{l}{\includegraphics[height=1em]{figures/hera.png} \textit{Ours} } \\
\midrule
\rowcolor{cyan!10} HERA (\qwen) & 55.38 & 52.50 & 63.03 & 60.02 & 59.50 & 64.77 & 27.19 & 26.57 & 35.82 & 54.43 & 48.74 & 67.81 & 49.01 & 46.46 & 60.53 & 67.35 \\
\rowcolor{cyan!10} HERA (\llama) & 53.28 & 48.01 & 57.91 & 52.62 & 46.70 & 55.42 & 25.04 & 24.46 & 31.07 & 48.53 & 43.46 & 60.72 & 46.60 & 45.22 & 56.31 & 60.31 \\
\bottomrule
\end{tabular}}
\end{table}
\vspace{10pt}

Table~\ref{tab:part1-data-swapped} highlights the performance of \hero compared to direct and CoT inference without RAG. Across all datasets, \hero exhibits substantial gains over baseline LLMs, with improvements particularly pronounced on multi-hop and ambiguous QA tasks. Notably, HERA (\qwen) consistently outperforms HERA (\llama), reflecting its superior capability in complex reasoning scenarios. Across metrics, \hero achieves more than double the F1 scores of standard CoT baselines on several datasets, indicating that experience-guided orchestration and prompt evolution provide strong advantages even without retrieval augmentation. Gains are also robust across both in-domain and sparse or ambiguous queries, demonstrating that structured inter-agent coordination and semantic group-level reasoning significantly enhance reasoning fidelity over single-agent inference.

\subsection{Ablation Studies Using \llama as the Backbone for \hero}
\label{app: ablte_llama}
\begin{figure}[ht]
\centering
        \includegraphics[width=1\linewidth]{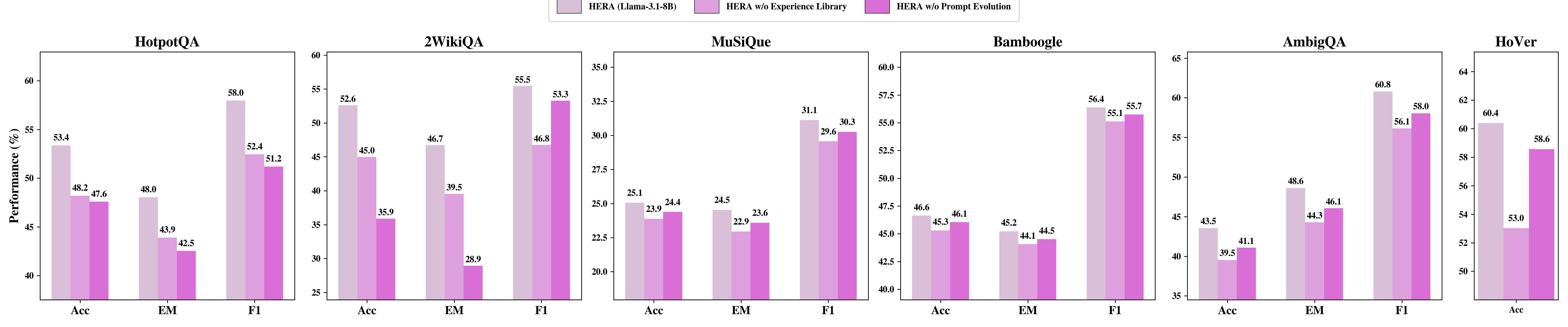}
        \caption{Ablation Studies of \hero with \llama as the backbone. }
        \label{fig:abla_llama}
\end{figure}

As shown in Figure~\ref{fig:abla_llama}, 
\hero with \llama as bacnkbone demonstrate similar trends with \hero-Qwen, exceot MusiQue and Bamboogle. Prompt Evolution is the most critical component. Its removal results in the largest drops across nearly all datasets. For example, F1 decreases from $58.0 \rightarrow 51.2$ on HotpotQA, $55.5 \rightarrow 53.3$ on 2WikiQA, and $60.8 \rightarrow 58.0$ on AmbigQA. This indicates that adaptive updating of agent prompts is essential for improving reasoning performance, particularly on multi-hop and ambiguous queries. 

Experience Library removal produces moderate but consistent declines, \eg, F1 drops from $58.0 \rightarrow 52.5$ on HotpotQA and $55.5 \rightarrow 46.75$ on 2WikiQA. This shows that stored experience contribute to performance stability and cross-instance generalization.

\section{Pseudocode}
\label{app:pseducode}

The pseudocode for the proposed \hero{} is presented below.

\begin{algorithm}[H]
\caption{\hero — Top Level}
\begin{algorithmic}[1]
\Require query $q$, corpus $\mathcal{D}$, orchestrator $\pi_{\mathcal{O}}$, iterations $T$
\Ensure optimized experience library $\mathcal{E}$, evolved agent prompts $\{\rho_1, \dots, \rho_K\}$

\State $\mathcal{E} \leftarrow \emptyset$
\State $\mathcal{N} \leftarrow \textsc{InitializeAgents}()$
\Comment{each $\mathcal{N}_i = (\pi_i, \rho_i, \mathcal{T}_i)$}

\For{$t = 1$ \textbf{to} $T$}
    \State Sample query $q \sim \mathcal{Q}$

    \Comment{Orchestration }
    \State $\Gamma \leftarrow \textsc{Orchestrate}(q, \mathcal{E}, \mathcal{N})$
    \Comment{sample topology: order + dependencies}
    \State $\tau \leftarrow \textsc{Execute}(\Gamma, q, \mathcal{D})$
    \Comment{run agents, produce trajectory}

    \Comment{Gradient-free GRPO-style orchestrator update}
    \State $\textsc{OrchestratorUpdate}(q, \mathcal{E}, \mathcal{N})$

    \Comment{RoPE: agent prompt evolution}
    \State $\mathcal{N}_{\text{fail}} \leftarrow \pi_{\mathcal{O}}.\textsc{IdentifyFailedAgents}(\tau)$
    \For{each $\mathcal{N}_i \in \mathcal{N}_{\text{fail}}$}
        \State $\textsc{RoPE}(\mathcal{N}_i, \tau)$
    \EndFor
\EndFor

\State \Return $\mathcal{E},\ \{\rho_1, \dots, \rho_K\}$
\end{algorithmic}
\end{algorithm}

\begin{algorithm}[H]
\caption{OrchestratorUpdate (GRPO-style)}
\begin{algorithmic}[1]
\Require query $q$, experience library $\mathcal{E}$, agent set $\mathcal{N}$, group size $G$

\State $\text{group} \leftarrow [\,]$
\For{$i = 1$ to $G$}
    \State $\Gamma_i \leftarrow \textsc{SampleTopology}(q, \mathcal{E}, \mathcal{N})$
    \State $\tau_i \leftarrow \textsc{Execute}(\Gamma_i, q, \mathcal{D})$
    \State $r_{\text{task}} \leftarrow \textsc{Evaluate}(\tau_i,\ \text{metric}=\text{F1})$
    \State $r_{\text{eff}} \leftarrow \textsc{Evaluate}(\tau_i,\ \text{metric}=\text{token\_cost})$
    \State $\text{group}.\textsc{Append}\bigl((\Gamma_i,\, \tau_i,\, r_{\text{task}},\, r_{\text{eff}})\bigr)$
\EndFor

\If{$\neg\,\textsc{HasMixedOutcomes}(\text{group})$}
    \State \Return
    \Comment{Reflection and insight extraction is only based on groups that have explicit successful and failed trajectories}
\EndIf

\State $\text{group}_{\text{sorted}} \leftarrow \textsc{Sort}\!\left(\text{group},\ \text{key}=(r_{\text{task}}{\downarrow},\, r_{\text{eff}}{\uparrow})\right)$
\Comment{task perf first, efficiency second}

\State $\mathcal{I} \leftarrow \pi_{\mathcal{O}}.\textsc{ReflectOnGroup}(\text{group}_{\text{sorted}})$
\Comment{natural language semantic advantages}

\State $\textsc{ExperienceLibraryUpdate}(\mathcal{E}, \mathcal{I}, q)$
\end{algorithmic}
\end{algorithm}

\begin{algorithm}[H]
\caption{ExperienceLibraryUpdate (ADD / MERGE / PRUNE / KEEP)}
\begin{algorithmic}[1]
\Require experience library $\mathcal{E}$, new insights $\mathcal{I}$, query $q$

\State $c \leftarrow \textsc{CharacterizeQuery}(q)$

\For{each insight $z \in \mathcal{I}$}
    \State $\text{matches} \leftarrow \textsc{Retrieve}(\mathcal{E}, z, c)$

    \If{$\text{matches} = \emptyset$}
        \State $\mathcal{E}.\textcolor{blue}{\textsc{Add}}\bigl((c,\, z,\, u{=}0)\bigr)$
        \Comment{novel insight}

    \ElsIf{$\textsc{Complementary}(z,\, \text{matches})$}
        \State $\mathcal{E}.\textcolor{orange}{\textsc{Merge}}(z,\, \text{matches})$
        \Comment{combine semantically similar entries}

    \ElsIf{$\textsc{Conflicts}(z,\, \text{matches})$}
        \State $\text{low}_u \leftarrow \textsc{FilterLowUtility}(\text{matches})$
        \State $\mathcal{E}.\textcolor{green}{\textsc{Prune}}(\text{low}_u)$
        \Comment{remove conflicting or stale entries}

    \Else
        \State $\mathcal{E}.\textcolor{purple}{\textsc{Keep}}()$
        \Comment{no change needed}
    \EndIf
\EndFor

\For{each $(c,\, z,\, u) \in \mathcal{E}$}
    \If{$z$ was used \textbf{and} outcome $=$ success}
        \State $u \leftarrow u + 1$
    \EndIf
\EndFor
\end{algorithmic}
\end{algorithm}

\begin{algorithm}[H]
\caption{Orchestrate (experience-guided topology sampling)}
\begin{algorithmic}[1]
\Require query $q$, experience library $\mathcal{E}$, agent set $\mathcal{N}$

\State $c \leftarrow \textsc{CharacterizeQuery}(q)$

\State $\mathcal{E}_{\text{rel}} \leftarrow [\,]$
\For{each $(c',\, z,\, u) \in \mathcal{E}$ sorted by $u$ descending}
    \If{$\textsc{Similar}(c', c)$ \textbf{and} $\neg\,\textsc{Redundant}(z,\, \mathcal{E}_{\text{rel}})$}
        \State $\mathcal{E}_{\text{rel}}.\textsc{Append}((z, u))$
        \Comment{balance utility and diversity}
    \EndIf
\EndFor

\State $\Gamma \sim \pi_{\mathcal{O}}\!\left(\cdot \mid q,\, \mathcal{E}_{\text{rel}},\, \mathcal{N}\right)$
\Comment{$\Gamma$ specifies agents, execution order, dependencies}

\State \Return $\Gamma$
\end{algorithmic}
\end{algorithm}

\begin{algorithm}[H]
\caption{RoPE — Role-aware Prompt Evolution}
\begin{algorithmic}[1]
\Require failed agent $\mathcal{N}_i$, global trajectory $\tau$

\State $\mathcal{B}_i \leftarrow \textsc{GetRecentFailures}(\mathcal{N}_i)$
\Comment{buffer of recent failed trajectories}

\State $\text{variants} \leftarrow [\,]$
\For{each $\text{axis} \in \{\text{efficiency},\ \text{thoroughness},\ \text{risk\_sensitivity}\}$}
    \State $\rho^{\text{var}} \leftarrow \textsc{GenerateVariant}(\rho_i,\ \text{axis})$
    \State $\text{variants}.\textsc{Append}(\rho^{\text{var}})$
\EndFor

\State $\text{results} \leftarrow [\,]$
\For{each $\rho^{\text{var}} \in \text{variants}$}
    \State $\mathcal{N}_i.\rho \leftarrow \rho^{\text{var}}$
    \State $\tau^{\text{var}} \leftarrow \textsc{ReExecute}(\tau,\, \mathcal{N}_i)$
    \Comment{replay full trajectory with variant prompt}
    \State $\text{results}.\textsc{Append}\bigl((\rho^{\text{var}},\, \tau^{\text{var}},\, \textsc{Eval}(\tau^{\text{var}}))\bigr)$
\EndFor

\State $\Delta\rho_i^{op} \leftarrow \textsc{ExtractOperationalRules}(\mathcal{B}_i,\ \text{results})$
\Comment{short-term: correct immediate failure patterns}
\State $\Delta\rho_i^{bp} \leftarrow \textsc{ExtractBehavioralPrinciples}(\mathcal{B}_i,\ \text{results})$
\Comment{long-term: generalize across failure patterns}

\State $\Delta\rho_i \leftarrow \Delta\rho_i^{op} + \Delta\rho_i^{bp}$

\State $\rho_i^{t+1} \leftarrow \Pi_{\mathcal{C}}\!\left(\rho_i^{t} \oplus \Delta\rho_i\right)$
\Comment{project onto length and coherence constraints $\mathcal{C}$}
\end{algorithmic}
\end{algorithm}

\begin{algorithm}[H]
\caption{TopologyMutation (structural fallback)}
\begin{algorithmic}[1]
\Require current topology $\Gamma$, failed agents $\mathcal{N}_{\text{fail}}$, agent pool $\mathcal{N}$
\Comment{triggered when trajectories consistently fail, \eg F1 $= 0$}

\State $\mathcal{C}_{\Gamma} \leftarrow [\,]$
\Comment{candidate topologies}

\For{each $\mathcal{N}_i \in \mathcal{N}_{\text{fail}}$}

    \State \textbf{// Option A: replace failed agent}
    \State $\mathcal{N}_i' \leftarrow \textsc{SelectAlternative}(\mathcal{N},\ \text{exclude}=\mathcal{N}_i)$
    \State $\Gamma_{\text{replace}} \leftarrow \Gamma\ \text{with}\ \mathcal{N}_i\ \text{replaced by}\ \mathcal{N}_i'$
    \State $\mathcal{C}_{\Gamma}.\textsc{Append}(\Gamma_{\text{replace}})$

    \State \textbf{// Option B: augment topology}
    \State $\mathcal{N}_i^{\text{new}} \leftarrow \textsc{CreateNewAgent}()$
    \State $\Gamma_{\text{augment}} \leftarrow \Gamma\ \text{with}\ \mathcal{N}_i^{\text{new}}\ \text{inserted after}\ \mathcal{N}_i$
    \State $\mathcal{C}_{\Gamma}.\textsc{Append}(\Gamma_{\text{augment}})$

\EndFor

\For{each $\Gamma' \in \mathcal{C}_{\Gamma}$}
    \State $\textsc{OrchestratorUpdate}(q,\, \mathcal{E},\, \mathcal{N},\ \text{topology}{=}\Gamma')$
    \Comment{feed candidates back into GRPO loop}
\EndFor
\end{algorithmic}
\end{algorithm}

\section{Case Studies}
\label{app:case_studies}

We provide case studies in this section, for both success and failed cases. Note: the \includegraphics[height=2em]{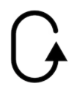} in every case figure means reuse the agent.
\subsection{Case 1 - Comparison Multi-hop QA}
\begin{figure}[ht!]
\centering
\includegraphics[width=0.5\linewidth]{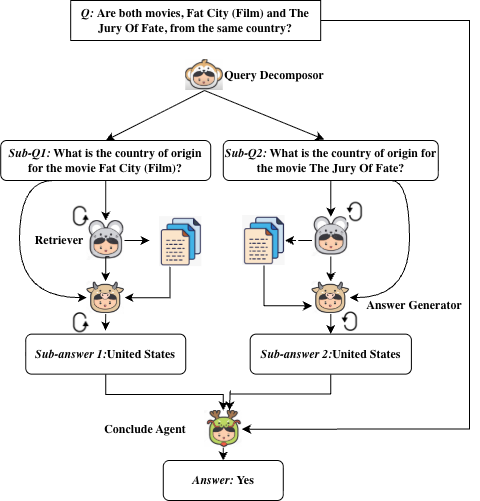}
    \caption{Case 1 - Comparison Multi-hop QA}
    \label{fig:case1}
\end{figure}

Case 1 (Fig. \ref{fig:case1}) exemplifies an optimal Comparison Multi-hop reasoning pattern: the orchestrator produces a parallel topology to independent entity facts (each movie’s country) are retrieved in parallel, followed by a serial aggregation step for comparison. Such a parallel-then-serial decomposition improves efficiency, minimizes redundancy, and scales naturally to multiple entities, demonstrating functional differentiation between knowledge grounding and reasoning control. 
\subsection{Case 2 - Causal Multi-hop QA}
\begin{figure}[ht!]
\centering
\includegraphics[width=0.6\linewidth]{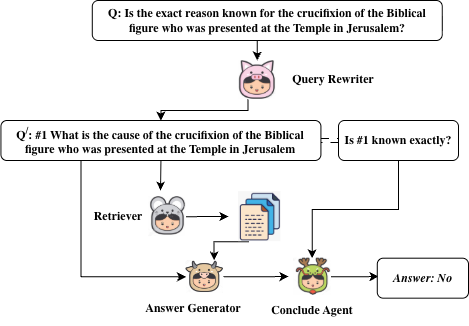}
    \caption{Case 2 - Causal Multi-hop QA}
    \label{fig:case2}
\end{figure}

For this causal multi-hop question (Fig. \ref{fig:case2}), \hero handles it by explicitly separating retrieval of causal evidence from reasoning over uncertainty. First, the system identifies the target entity (Jesus) and retrieves relevant historical and biblical sources describing the crucifixion and its contributing factors. Then, the reasoning module evaluates the retrieved evidence, accounting for ambiguity or contested interpretations, to determine whether the exact cause is definitively known. By maintaining this serial, dependency-aware workflow, \hero ensures that each reasoning hop builds on grounded evidence, minimizing hallucinations and producing a high-confidence answer. This functional differentiation between experience-grounded retrieval and adaptive causal reasoning allows HERA to robustly handle multi-hop questions where cause and certainty must be inferred sequentially.

\subsection{Case 3 - Temporal Multi-hop Q}
\begin{figure}[ht!]
\centering
\includegraphics[width=0.5\linewidth]{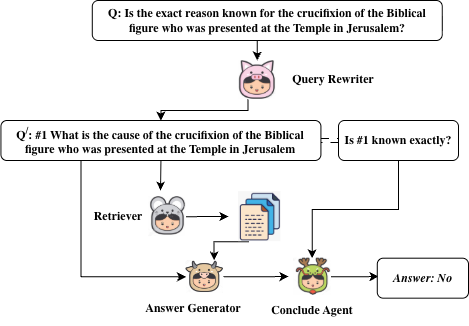}
    \caption{Case 3 - Temporal Multi-hop QA}
    \label{fig:case3}
\end{figure}
Case 3 (Fig.\ref{fig:case3}) requires computing the temporal intersection of two activity periods, which demands not just parallel retrieval but a subsequent arithmetic-logical operation on the retrieved intervals. The system correctly decomposes the query and retrieves accurate individual activity periods: 1914–1938 for Hutchison and 1921–1944 for Howard. However, the Answer Generator returns ``1914–1944" — the union of the two intervals rather than their intersection (which would correctly be 1921–1938).
This failure reveals a fundamental limitation: the system lacks an explicit set-theoretic reasoning module. This suggests that for questions requiring numerical or logical operations over retrieved facts, a dedicated symbolic computation step is necessary. The failure is not a retrieval failure but a reasoning composition failure — a distinction with important implications for system design.

\subsection{Case 4 - Intersection Multi-hop QA}

\begin{figure}[ht!]
\centering
\includegraphics[width=0.5\linewidth]{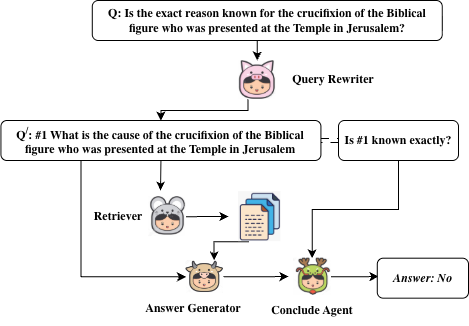}
    \caption{Case 4 - Intersection Multi-hop QA}
    \label{fig:case4}
\end{figure}
For Case 4 (Fig.\ref{fig:case4}), which is a Intersection Multi-Hop type, the error illustrates a critical challenge in handling multi-entity property overlap. The pipeline’s goal is to identify a property that applies to both Pavel Alexandrov and Valentin Turchin. Here, the intended property is ``Soviet". However, the Query Rewriter reformulated the query around ``what they were known for," shifting focus to professional achievements (mathematician, computer scientist) rather than shared nationality or affiliation. Consequently, the Retriever retrieved evidence on their careers, and the Evidence Selector highlighted disciplinary roles, leading the Conclude Agent to output ``scientists."

Insights from this case:
\begin{itemize}
    \item Intersection multi-hop questions require careful alignment of the property type, for example nationality, affiliation, or field must be explicitly targeted; otherwise, retrieval can be biased toward more salient but irrelevant attributes.
    \item Parallel entity retrieval is insufficient if the property semantics are ambiguous, as each agent may return correct facts individually, but the intersection reasoning fails if the property dimension differs.
    \item Conclude Agent reasoning is sensitive to initial decomposition: even small shifts in question framing (profession vs. nationality) propagate to a semantically plausible but incorrect intersection.
\end{itemize}

\end{document}